\documentclass[10pt,twocolumn,letterpaper]{article}

\PassOptionsToPackage{table}{xcolor}
\usepackage{cvpr}              
\definecolor{cvprblue}{rgb}{0.21,0.49,0.74}
\usepackage[pagebackref,breaklinks,colorlinks,allcolors=cvprblue]{hyperref}
\usepackage{xspace}
\usepackage{amsmath}
\usepackage{multirow}
\usepackage[accsupp]{axessibility}
\usepackage{bm}
\def\paperID{} 
\def\confName{CVPR}
\def\confYear{2026}

\usepackage{booktabs}
\usepackage{tabularx}
\definecolor{lightblue}{RGB}{224,241,255}
\usepackage{graphicx}

\title{AIGI-26: A Benchmark for AI-Generated Image Detection Across Diverse Modern Generators}

\author{
Qishuai Ding$^{1}$ \quad
Yifeng Wang$^{2}$ \quad
Shichuang Xie$^{1}$ \quad
Yanli Ren$^{1}$ \quad
Guorui Feng$^{1}$ \\
$^{1}$School of Communication and Information Engineering, Shanghai University, China \\
$^{2}$Tsinghua University, China \\
\texttt{Corresponding author: grfeng@shu.edu.cn}
}

\begin{document}
\maketitle

\begin{abstract}
Existing AI-generated image (AIGI) detection benchmarks predominantly rely on earlier synthesis systems with limited content diversity and conspicuous artifacts, failing to fully reflect the detection challenges posed by high-fidelity images from contemporary models. 
We introduce AIGI-26, a benchmark of 253,200 images covering 26 recent open- and closed-source generators. Its novel, diverse, and heterogeneous sources provide a basis for systematically evaluating detection methods on modern generated imagery.
Additionally, to capture subtle statistical discrepancies across complementary forensic views, we propose the multi-aspect forensic detection network (MAFDNet).
Its fixed forensic operators expose complementary texture, local prediction, resampling, and chromatic responses, summarized by a hierarchical encoder using multiscale means and standard deviations.
In parallel, a spectral branch captures Fourier log-magnitude and neighboring-frequency phase variation across radial bands and directional axes.
Fusing these descriptors combines local inconsistencies with global spectral organization for authenticity detection.
Trained on 12,660 AIGI-26 images, MAFDNet achieves 88.5\% accuracy and 94.9\% average precision on the 240,540-image test set, with only 1.04M parameters and 1.39 GFLOPs per $256\times256$ input.
MAFDNet also demonstrates effective cross-dataset generalization on DailyBench and Treasure-37.
The dataset and model are available at \textcolor[RGB]{90,155,210}{https://github.com/Dingding250/AIGI-26.}
\end{abstract}

\begin{figure*}[t]
    \centering
    \includegraphics[width=0.97\textwidth]{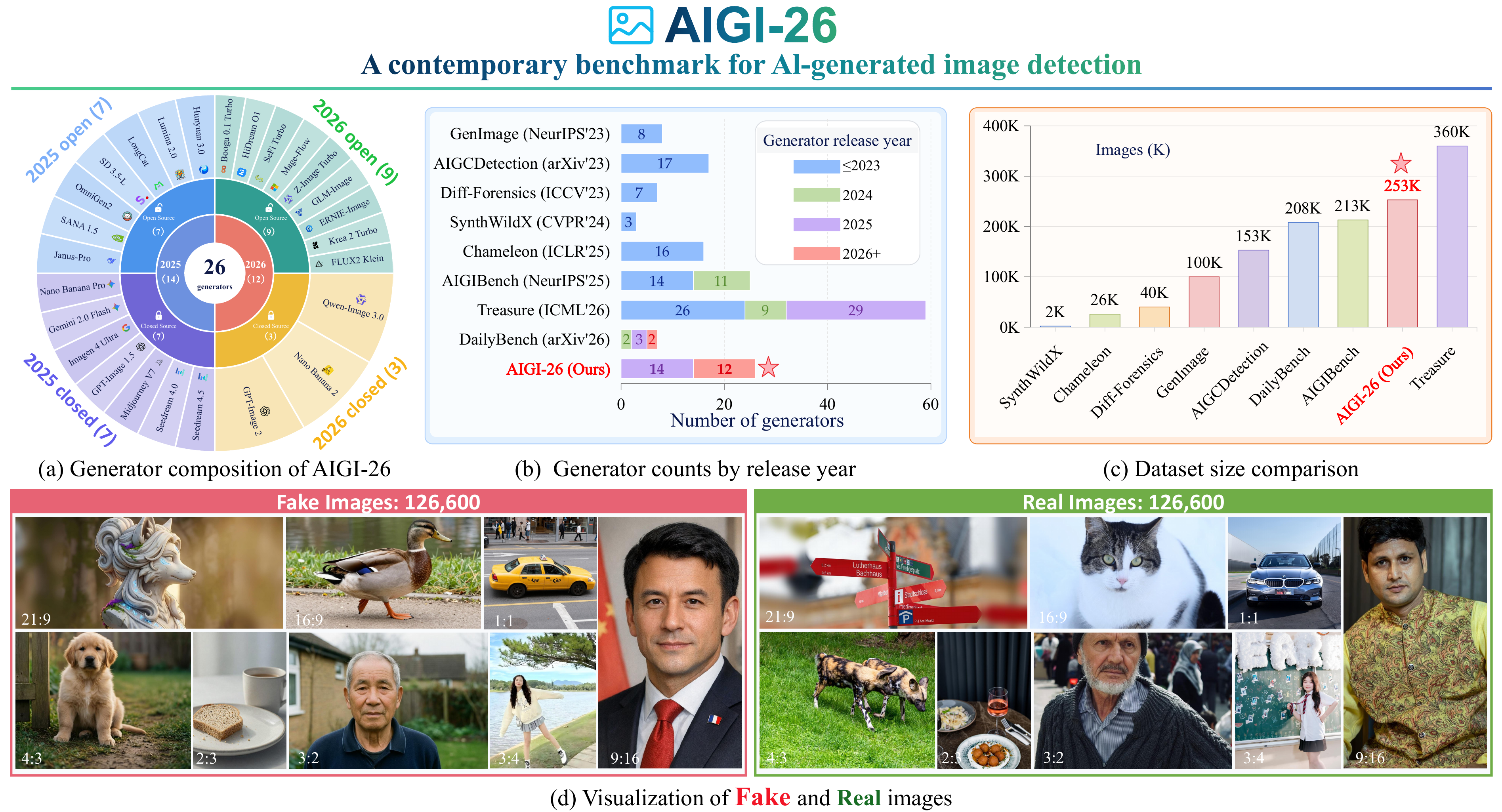}
    \caption{Overview of AIGI-26. (a) Its 26 generators by release year and accessibility. SD 3.5-L and Z-Image Turbo use 2025 and 2026 versions, respectively. Generator sources are detailed in the supplementary material. (b) Generator coverage by release year. (c) Image counts for general-purpose AIGI detection. GenImage~\cite{Zhu_2023_GenImage} denotes its standard test split. (d) Generated (fake) and real examples.}
    \label{fig:aigi26_dataset}
\end{figure*}

\section{Introduction}
\label{sec:intro}

Recent advances in text-to-image (T2I) generation increasingly blur the visual distinction between real and generated images, challenging visual evidence and public trust~\cite{Yan_2025_AIDE,Li_2026_IAPL}.
As generation pipelines evolve, their artifacts change and become less conspicuous, making generated images harder to detect~\cite{Wang_2026_Fleet}.
Reliable AI-generated image (AIGI) detection requires evaluation data that track contemporary systems and representations that capture forensic evidence across heterogeneous sources~\cite{Guillaro_2025_BFree,Qin_2026_GAPL}.

Early detection benchmarks primarily cover generative adversarial networks (GANs) and early diffusion models~\cite{Wang_2020_CNNDetection,Zhu_2023_GenImage}, whose outputs often exhibit lower visual fidelity and more conspicuous artifacts than those of recent systems~\cite{Yan_2025_AIDE,Jiang_2026_DailyBench}.
AIGIBench~\cite{Li_2025_AIGIBench} and Chameleon~\cite{Yan_2025_AIDE} expand coverage to more advanced generators and perceptually challenging images.
Treasure~\cite{Wang_2026_Fleet} and DailyBench~\cite{Jiang_2026_DailyBench} further incorporate high-fidelity T2I systems, including image generation supported by multimodal large language models (MLLMs).
Yet increasingly capable generators continue to emerge, and detection performance on their evolving output distributions remains insufficiently characterized by existing evaluations.

Closing this gap requires contemporary images from multiple generation sources, so that detection methods can be examined beyond the artifacts of any one model or generation family.
Therefore, we introduce AIGI-26, a dataset of 253,200 images comprising 126,600 generated images and 126,600 real images.
Its 26 recent generators include 16 open-source and 10 closed-source systems (Fig.~\ref{fig:aigi26_dataset}(a)), providing diverse, high-quality images from multiple generation sources.
Compared with existing benchmarks, AIGI-26 offers broader coverage of recent generative models, particularly those released in 2026, encompassing 14 generators from 2025 and 12 from 2026 onward (Fig.~\ref{fig:aigi26_dataset}(b)).
Moreover, AIGI-26 combines recent generator coverage with a large image collection for comprehensive detection evaluation (Fig.~\ref{fig:aigi26_dataset}(c)). This enables contemporary multi-source evaluation beyond the artifacts of earlier generators.

On the other hand, detecting modern AIGIs requires sensitivity to subtle artifacts across heterogeneous generators. Existing detectors leverage forensic cues such as gradients, local residuals, and pixel dependencies to expose synthesis-related irregularities~\cite{Tan_2023_LGrad,Tan_2024_NPR,Zhou2026BSF}. However, as generative pipelines diversify, the strength and distribution of these cues may change, limiting the reliability of individual forensic features. Meanwhile, frequency-based methods such as FreqNet~\cite{Tan_2024_FreqNet} and DFFreq~\cite{Yan_2026_DFFreq} exploit amplitude, phase, and subband information to reveal synthesis traces. Nevertheless, global spectral aggregation may obscure weak discrepancies confined to specific frequency ranges or orientations. The challenge is to preserve and combine these cues without suppressing weak artifacts. This motivates a compact detector that integrates multiscale spatial forensic statistics with structured spectral representations for detecting artifacts across diverse generators.

To realize this design, we propose the multi-aspect forensic detection network (MAFDNet).
Fixed operators emphasize subtle texture, pixel-dependency, and color discrepancies that can distinguish generated images from real ones.
A spatial encoder summarizes their strength and variability across scales to capture differences in fine detail and broader spatial patterns.
In parallel, Fourier magnitude and neighboring-frequency phase statistics target artifacts in component strength and phase organization.
Radial and directional pooling separate these cues by frequency scale and orientation, allowing the detector to exploit discrepancies that whole-spectrum averaging can conceal.
Fusing both branches combines local artifact evidence with global spectral differences for authenticity classification.

Our contributions are threefold:
\begin{itemize}
    \item We introduce AIGI-26, a balanced benchmark of 253,200 images from 26 recent open- and closed-source generators, providing novel, diverse, and high-quality detection samples from multiple modern generation sources.
    \item We propose MAFDNet, a multi-aspect detector that combines fixed forensic views with multiscale spatial moments and radial--directional spectral statistics, capturing complementary local and global forensic evidence.
    \item On AIGI-26, MAFDNet achieves 88.5\% Acc and 94.9\% AP with 1.04M parameters and 1.39 GFLOPs. It also delivers competitive performance on DailyBench and Treasure-37, with unseen-generator results further demonstrating generalization beyond the training sources.
\end{itemize}

\section{Related Work}
\label{sec:related}

\subsection{Benchmarks and Evaluation}

GenImage~\cite{Zhu_2023_GenImage} and DiffusionForensics~\cite{Wang_2023_DIRE} support evaluation across GAN and diffusion synthesis, while Chameleon~\cite{Yan_2025_AIDE} emphasizes perceptually convincing images.
Fake2M~\cite{Lu_2023_MPBench} supports human and model perception studies, and Community Forensics~\cite{Park_2025_CommunityForensics} extends community-model coverage.
WildFake~\cite{Hong_2025_WildFake} broadens source and style diversity, and AIGIBench~\cite{Li_2025_AIGIBench} examines multiple generation pipelines, degradations, and preprocessing.
The 2026 benchmarks Treasure~\cite{Wang_2026_Fleet} and DailyBench~\cite{Jiang_2026_DailyBench} further advance evaluation: Treasure spans heterogeneous architectures and commercial systems, while DailyBench emphasizes high visual quality and covers both full-image generation and localized manipulation.
However, broad coverage does not ensure assessment of the latest generators.
Benchmark scores reflect the generator versions available at dataset construction.
AIGI-26 focuses on modern generated images. Figs.~\ref{fig:aigi26_dataset}(b) and~\ref{fig:aigi26_dataset}(c) compare generator release-year coverage and dataset size.

\subsection{Spatial and Spectral Forensic Evidence}

Spatial methods examine texture, gradients, neighboring-pixel correlations, and color or noise statistics~\cite{Liu_2020_GramNet,Tan_2023_LGrad,Tan_2024_NPR,Jia_2025_Color,Zhang_2026_FALCON}.
GramNet~\cite{Liu_2020_GramNet} models global texture to improve transfer across generators and robustness to degradations such as JPEG compression, blur, and downsampling.
NPR~\cite{Tan_2024_NPR} captures local pixel dependencies induced by upsampling in GAN and diffusion generators.
BSF~\cite{Zhou2026BSF} uses pixel-level mapping to suppress semantic shortcuts and emphasize high-frequency artifacts.
Spectral methods characterize frequency distributions and amplitude or phase cues~\cite{Frank_2020_Frequency,Durall_2020_Spectral,Tan_2024_FreqNet}.
DFFreq~\cite{Yan_2026_DFFreq} combines wavelet and Fourier features, while PRISM~\cite{Ricco2025PRISM} pools radial magnitude and phase statistics.
Reconstruction-based methods expose discrepancies through diffusion inversion or reconstruction~\cite{Wang_2023_DIRE,Cazenavette_2024_FakeInversion,Luo_2024_LaRE2}.
LTD~\cite{Yang_2026_LTD} captures latent transition discrepancies across selected intermediate layers to distinguish real and generated images.
MAFDNet combines fixed forensic responses, spatial moments, and radial--directional spectral statistics to capture complementary local and global artifacts.

\section{Construction of AIGI-26}
\label{sec:data}

\subsection{Image Acquisition and Curation}
\label{sec:data_sources}

\begin{figure}[t]
    \centering
    \includegraphics[width=\linewidth]{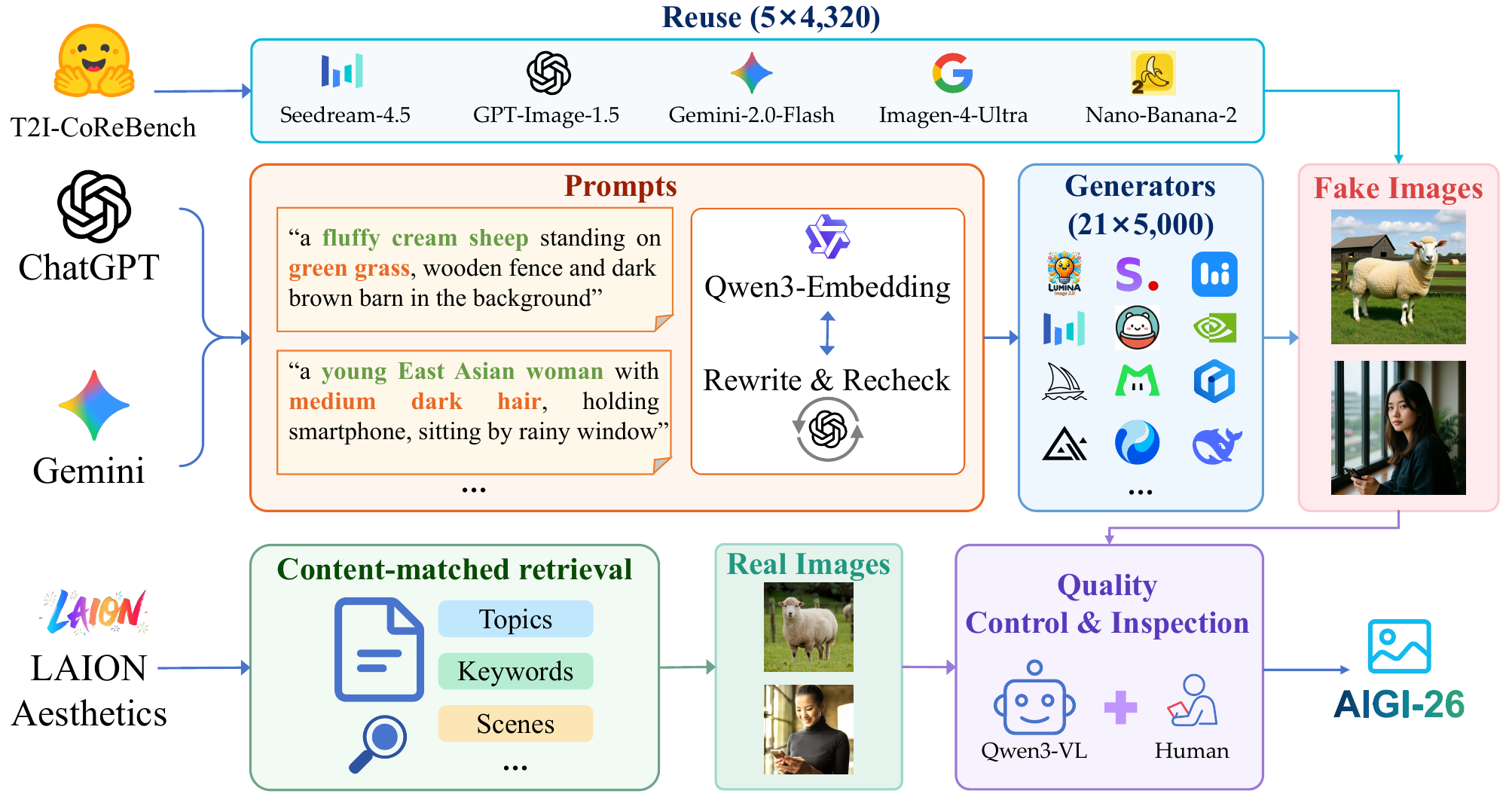}
    \caption{Construction pipeline of AIGI-26.}
    \label{fig:data_construction}
\end{figure}

The 26 generators in Fig.~\ref{fig:aigi26_dataset}(a) supply locally generated, API-generated, and reused benchmark images. Fig.~\ref{fig:data_construction} outlines acquisition and curation.
ChatGPT and Gemini produce 3,000 and 2,000 seed prompts, respectively, expanded through subject recombination and sentence transformations.
After semantic curation (Sec.~\ref{sec:data_quality}), prompts are randomly grouped and assigned to generators.
Fifteen local models and six API-accessed models each contribute 5,000 images.
To reduce collection costs, we reuse 4,320 outputs from each of five generators in T2I-CoReBench~\cite{Li2026T2ICoReBench} for authenticity detection.
T2I-CoReBench uses these outputs to evaluate compositional and reasoning capabilities in T2I generation.
Prompts from the 21 directly accessed generators are matched to the \texttt{text} field of LAION-Aesthetics V2~\cite{LAION_Aesthetics_V2}. For the five reused subsets, Qwen3-VL-4B-Instruct~\cite{Qwen3VL4B_Model} first produces image descriptions for the same matching procedure. Matching shared topics, keywords, and scenes aligns image content across classes, yielding 126,600 real images paired with the 126,600 generated images.

\subsection{Coverage and Evaluation Split}
\label{sec:data_coverage}
\label{sec:data_protocol}

The 253,200 images cover people, animals, objects, landscapes, and other subjects (Fig.~\ref{fig:aigi26_dataset}(d)), with eight image-size settings spanning square, portrait, and landscape formats.
We construct a class-balanced 5\%/95\% training--testing split of 12,660/240,540 images.
Generated images sharing a prompt across generators stay in one partition, while real images are grouped by matched topics and scenes.
Duplicate image IDs are removed before partitioning to prevent reuse of the same indexed sample.
Group-level sampling is stratified by generator and authenticity class to preserve comparable class proportions across partitions.

\subsection{Quality Control}
\label{sec:data_quality}

We identify semantically duplicate prompts using Qwen3-Embedding-0.6B~\cite{Qwen3Embedding06B_Model} with a cosine-similarity threshold of 0.7. ChatGPT rewrites duplicate or unreasonable descriptions for rechecking.
Qwen3-VL-4B-Instruct~\cite{Qwen3VL4B_Model} screens both generated and retrieved real images for content completeness and visual quality, with flagged failures and retrieval mismatches reviewed by humans while generation artifacts are retained.
Topic/scene matching reduces class-dependent semantic shortcuts~\cite{Guillaro_2025_BFree}.
Image-ID checks prevent repeated real-image matching and duplicate entries before partitioning.
Together with prompt grouping, this prevents reuse of indexed samples across partitions and reduces leakage from repeated content or scenes.
Prompt processing, screening, image-ID checks, and model settings are detailed in the supplementary material.

\begin{figure*}[t]
    \centering
    \includegraphics[width=\textwidth]{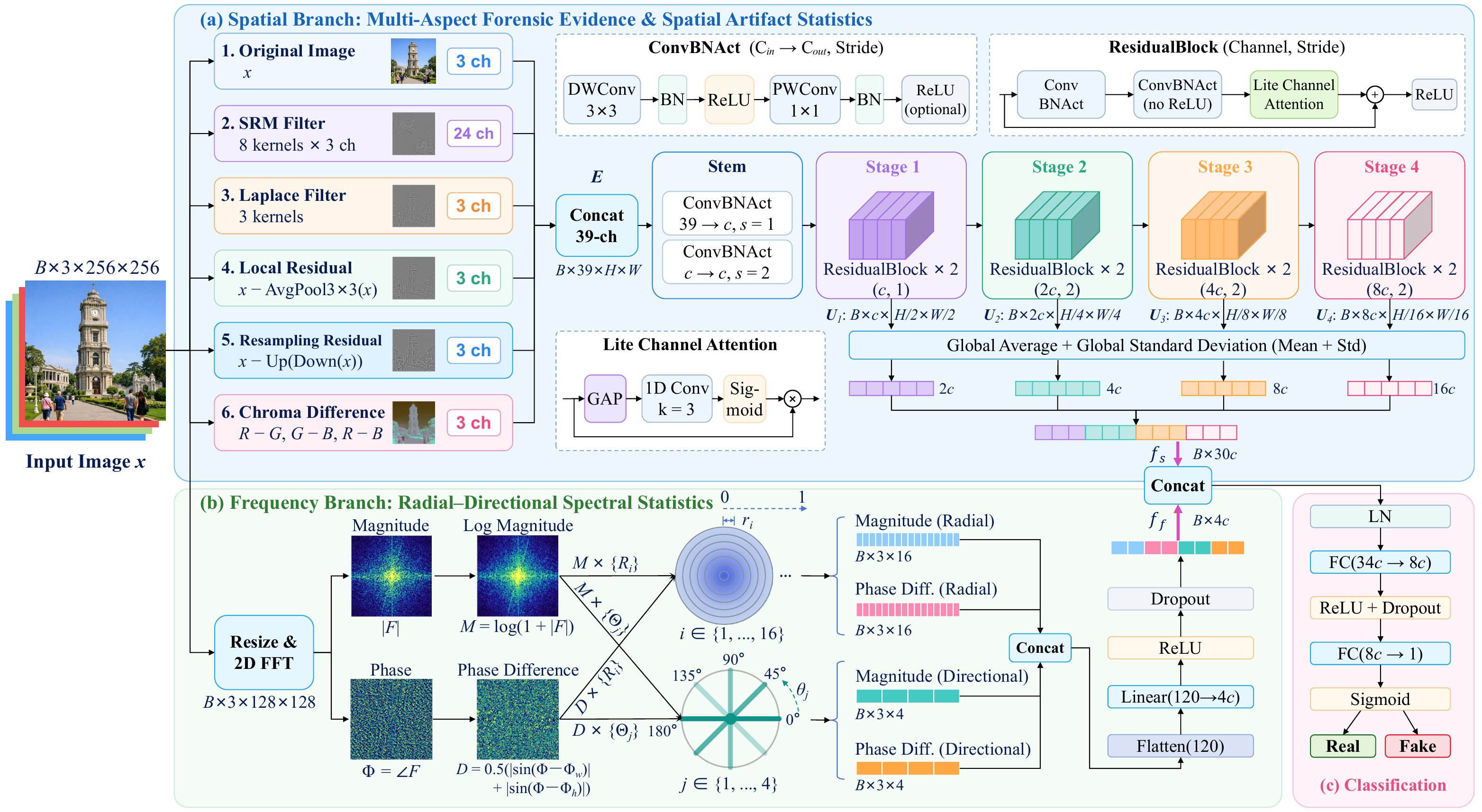}
    \caption{Overview of MAFDNet. (a) The spatial branch combines a fixed forensic evidence bank with a hierarchical encoder and multiscale mean/standard-deviation pooling. (b) The spectral branch aggregates Fourier log-magnitude and neighboring-frequency phase variation with radial and directional masks. (c) The fused descriptors support binary authenticity classification.}
    \label{fig:mafdnet}
\end{figure*}

\section{Method}
\label{sec:method}

Our goal is to distinguish real and generated images through subtle texture, color, and frequency discrepancies beneath plausible content.
MAFDNet (Fig.~\ref{fig:mafdnet}) combines spatial evidence that highlights local irregularities with spectral statistics that capture differences in image-wide frequency organization.
Given an RGB image $x\in\mathbb{R}^{3\times H\times W}$, the two branches yield spatial and spectral descriptors $f_s$ and $f_f$.

\subsection{Multi-Aspect Forensic Evidence}
\label{sec:mafd_evidence}

To expose weak artifacts beneath scene content, we emphasize texture, pixel-dependency, and color differences between real and generated images through six views: original image, spatial rich model (SRM) filter~\cite{Fridrich_2012_SRM}, Laplace filter, local residual, resampling residual, and chroma difference (Fig.~\ref{fig:mafdnet}(a)).
The SRM filter bank $\mathcal{S}$ applies eight fixed $5\times5$ high-pass kernels per RGB channel. The Laplace filter bank $\mathcal{L}$ applies three fixed $3\times3$ kernels to $\bar{x}=(x_R+x_G+x_B)/3$.
Their responses highlight fine texture and intensity transitions where generation artifacts may occur.
Each fixed kernel is divided by its absolute coefficient sum to limit scale differences caused by the filters themselves.
The remaining cues are
\begin{equation}
\begin{aligned}
 r_l &= x-\operatorname{Avg}_{3\times3}(x),\\
 r_u &= x-\operatorname{Up}(\operatorname{Down}_{1/2}(x)),\\
 r_c &= [x_R-x_G;\,x_G-x_B;\,x_R-x_B].
\end{aligned}
\label{eq:mafd_residuals}
\end{equation}
Here $\operatorname{Avg}_{3\times3}$ is local averaging. Bilinear downsampling halves the resolution and upsampling restores the original size.
The residuals $r_l$ and $r_u$ expose local prediction errors and resampling-sensitive detail, helping the encoder identify atypical pixel relationships.
Chroma differences $r_c$ expose color-channel relationships for detecting chromatic inconsistencies, while RGB retains appearance context.
We concatenate the six groups as
\begin{equation}
\begin{split}
 E=[x;\,&\tanh(3\mathcal{S}(x));\,\tanh(3\mathcal{L}(\bar{x}));\\
       &\tanh(4r_l);\,\tanh(4r_u);\,r_c]
       \in\mathbb{R}^{39\times H\times W},
\end{split}
\label{eq:mafd_evidence}
\end{equation}
where $[\,;\,]$ denotes concatenation.
The tanh transforms limit dominance by large responses while preserving their signs.

\subsection{Spatial Artifact Statistics}
\label{sec:mafd_spatial}

Generation artifacts may affect both fine textures and broader spatial patterns. A hierarchical residual encoder processes $E$ with increasing receptive fields to distinguish real and generated images at both scales.
A two-layer stem uses strides $1$ and $2$, followed by four stages with channel widths $(c,2c,4c,8c)$.
Each stage contains two residual blocks, with first-block strides of $(1,2,2,2)$.
Depthwise $3\times3$ convolutions model local patterns, and pointwise $1\times1$ convolutions mix the forensic responses across channels.
To emphasize responses useful for authenticity discrimination, lite channel attention (LCA) reweights residual features before shortcut addition. Following ECA~\cite{Wang_2020_CVPR}, it uses global average pooling, a 1D channel convolution of kernel size three, and sigmoid gating.
We set $c=40$ and examine other widths in the ablation study.

Artifacts can differ in both response strength and spatial variability. We retain both by concatenating channel-wise spatial means and population standard deviations of stage outputs $U_1,\ldots,U_4$:
\begin{equation}
 f_s=\mathop{\operatorname{Cat}}_{t=1}^{4}
       [\operatorname{Mean}(U_t);\operatorname{Std}(U_t)]
       \in\mathbb{R}^{30c}.
\label{eq:mafd_spatial}
\end{equation}
These moments summarize potential artifact responses across scales without requiring a fixed spatial location.

\subsection{Radial--Directional Spectral Statistics}
\label{sec:mafd_spectral}

This branch targets spectral artifacts that can distinguish generated images from real ones beyond local detail. Magnitude statistics capture differences in frequency-component strength, while neighboring-frequency phase variation captures changes in phase organization.
Inputs whose longer side exceeds $128$ pixels are bilinearly downsampled with the aspect ratio preserved.
Let $F$ denote the channel-wise orthonormal 2D Fourier transform of the branch input. Its phase is $\Phi=\arg F$.
We compute log-magnitude $M$ and a phase-variation map $D$:
\begin{equation}
\begin{aligned}
 M&=\log(1+|F|),\\
 D&=\tfrac12\bigl(|\sin(\Phi-\Phi_w)|+|\sin(\Phi-\Phi_h)|\bigr),
\end{aligned}
\label{eq:mafd_spectrum}
\end{equation}
where $\Phi_w$ and $\Phi_h$ are one-bin circular shifts along the horizontal and vertical frequency axes.
Log compression reduces magnitude dominance to retain weaker spectral cues. The absolute sine maps neighboring-phase differences to bounded responses, averaged over both axes to summarize local phase variation continuously across phase wraps.

The 16 radial masks $R_i$ isolate magnitude and phase differences between real and generated images within frequency bands. Four directional masks $\Theta_j$ emphasize orientation-dependent discrepancies otherwise mixed by radial averaging.
For frequency position $q$, $r(q)\in[0,1]$ is its distance from the frequency origin divided by the maximum grid radius, and $\theta(q)\in[0,2\pi)$ specifies its direction, with $\theta(0)=0$.
The coordinates are defined on the signed frequency grid, with masks aligned to the spectral maps.
For $i=1,\ldots,16$ and $j=1,\ldots,4$, the masks are
\begin{equation}
\begin{aligned}
 R_i(q)&=\mathbf{1}\!\left[\frac{i-1}{16}\leq r(q)<\frac{i}{16}\right],\\
 \Theta_j(q)&=\cos^2\!\left(\theta(q)-\frac{(j-1)\pi}{4}\right).
\end{aligned}
\label{eq:mafd_masks}
\end{equation}
where $\mathbf{1}[\cdot]$ is the indicator function, and the last radial mask also includes $r=1$.
The squared cosine smoothly emphasizes frequencies aligned with the $0^\circ$, $45^\circ$, $90^\circ$, and $135^\circ$ axes using nonnegative weights equal for opposite directions.

Within each RGB channel, we average $M$ and $D$ to compare log-magnitude and phase variation across bands and directions, using $w_A(q)=A(q)/\sum_{q'}A(q')$ for $A\in\{R_i,\Theta_j\}$ (zero for an empty mask):
\begin{equation}
\begin{aligned}
 M_R&=\Bigl[\textstyle\sum_q w_{R_i}(q)M(q)\Bigr]_{i=1}^{16},\\
 D_R&=\Bigl[\textstyle\sum_q w_{R_i}(q)D(q)\Bigr]_{i=1}^{16},\\
 M_{\Theta}&=\Bigl[\textstyle\sum_q w_{\Theta_j}(q)M(q)\Bigr]_{j=1}^{4},\\
 D_{\Theta}&=\Bigl[\textstyle\sum_q w_{\Theta_j}(q)D(q)\Bigr]_{j=1}^{4},
\end{aligned}
\label{eq:mafd_weighted_stats}
\end{equation}
where $M_R,D_R$ average log-magnitude and phase variation within each radial band, while $M_{\Theta},D_{\Theta}$ pool the same maps with four axial weightings, retaining band-specific and directional artifact cues.

Columns follow mask order, yielding $M_R,D_R\in\mathbb{R}^{3\times16}$ and $M_{\Theta},D_{\Theta}\in\mathbb{R}^{3\times4}$.
Concatenation and projection give
\begin{equation}
\begin{aligned}
 t&=\operatorname{Cat}[M_R;D_R;M_{\Theta};D_{\Theta}]
       \in\mathbb{R}^{3\times40},\\
 f_f&=\operatorname{Proj}\!\left(\operatorname{Flatten}(t)\right)
       \in\mathbb{R}^{4c}.
\end{aligned}
\label{eq:mafd_spectral_concat}
\end{equation}
We flatten the 40 statistics per RGB channel into a 120-dimensional representation and project it to $4c$ dimensions using a linear layer, ReLU, and dropout (Fig.~\ref{fig:mafdnet}(b)).

\subsection{Fusion and Detection}
\label{sec:mafd_fusion}

We concatenate the complementary descriptors and predict the generated-image probability:
\begin{equation}
 p=\operatorname{Sigmoid}\!\left(\operatorname{MLP}([f_s;f_f])\right).
\label{eq:mafd_prediction}
\end{equation}
The classifier applies layer normalization followed by an MLP of widths $34c\rightarrow8c\rightarrow1$, with intermediate ReLU and dropout (Fig.~\ref{fig:mafdnet}(c)).
For $N$ labeled images, binary supervision is formulated as
\begin{equation}
 \mathcal{L}_{\mathrm{cls}}=-\frac1N\sum_{n=1}^{N}
       [y_n\log p_n+(1-y_n)\log(1-p_n)],
\label{eq:mafd_loss}
\end{equation}
where $y_n=1$ denotes an AI-generated image and $p_n$ is its predicted probability.
The spatial encoder, spectral projection, and classifier are trained jointly.

\section{Experiments}
\label{sec:experiments}

\begin{table*}[t]
\caption{Detection on AIGI-26 before and after training on its shared split. Detection metrics are reported in \%. \textbf{Bold}/\underline{underline} mark the best/second-best scores.}
\label{tab1}
\centering
\setlength{\tabcolsep}{3pt}
\fontsize{8pt}{9.8pt}\selectfont
\begin{tabularx}{\textwidth}{@{}lccrr*{8}{>{\centering\arraybackslash}X}@{}}
\toprule
\multirow{2}{*}{Method}
& \multirow{2}{*}{Venue}
& \multirow{2}{*}{Input}
& \multirow{2}{*}{Params$\downarrow$}
& \multirow{2}{*}{FLOPs$\downarrow$}
& \multicolumn{4}{c}{\shortstack{Official Pretrained Model}}
& \multicolumn{4}{c}{Trained on AIGI-26} \\

\cmidrule(lr){6-9}
\cmidrule(lr){10-13}

& & & & &
Acc$\uparrow$ &   AP$\uparrow$ & R-Acc$\uparrow$ & F-Acc$\uparrow$
& Acc$\uparrow$ &   AP$\uparrow$ & R-Acc$\uparrow$ & F-Acc$\uparrow$ \\

\midrule

ResNet-50~\cite{He_2016_ResNet}
& CVPR'16
& $1\!\times\!3\!\times\!224^2$
& 1.43M & 2.29G
& 51.4 & 58.2 & 98.7 & 4.2
& 68.6 & 73.6 & 56.9 & 80.3 \\

CNND~\cite{Wang_2020_CNNDetection}
& CVPR'20
& $1\!\times\!3\!\times\!224^2$
& 23.51M & 4.13G
& 52.8 & \textbf{59.2} & 96.2 & 9.4
& 84.8 & 92.2 & 83.1 & 86.6 \\

GramNet~\cite{Liu_2020_GramNet}
& CVPR'20
& $1\!\times\!3\!\times\!224^2$
& 11.73M & 2.27G
& 52.3 & 56.3 & 92.7 & 11.9
& 67.0 & 72.3 & 52.3 & 81.8 \\

LGrad~\cite{Tan_2023_LGrad}
& CVPR'23
& $1\!\times\!3\!\times\!224^2$
& 23.51M & 4.13G
& 49.9 & 48.4 & 98.9 & 0.9
& 55.0 & 56.1 & 51.4 & 58.6 \\

UnivFD~\cite{Ojha_2023_UniversalFakeDetect}
& CVPR'23
& $1\!\times\!3\!\times\!224^2$
& 427.62M & 51.90G
& \underline{52.9} & 53.7 & 54.8 & \textbf{50.9}
& 78.5 & 85.7 & 77.9 & 79.2 \\

NPR~\cite{Tan_2024_NPR}
& CVPR'24
& $1\!\times\!3\!\times\!224^2$
& 1.44M & 1.75G
& 51.8 & 56.5 & 98.1 & 5.4
& 70.5 & 78.6 & 71.2 & 69.8 \\

LaDeDa~\cite{Cavia_2024_LaDeDa}
& arXiv'24
& $1\!\times\!3\!\times\!224^2$
& 13.64M & 16.13G
& 50.7 & \underline{58.6} & 99.6 & 1.7
& 83.1 & 90.1 & 85.2 & 81.0 \\

FreqNet~\cite{Tan_2024_FreqNet}
& AAAI'24
& $1\!\times\!3\!\times\!224^2$
& 1.85M & 1.97G
& \textbf{55.4} & 57.1 & 78.1 & 32.6
& 67.2 & 72.3 & 73.4 & 61.1 \\

AIDE~\cite{Yan_2025_AIDE}
& ICLR'25
& $5\!\times\!3\!\times\!256^2$
& 897.83M & 225.69G
& 49.7 & 51.0 & 99.3 & 0.1
& \underline{87.1} & \underline{94.5} & \underline{86.1} & \underline{88.2} \\

SAFE~\cite{Li_2025_SAFE}
& KDD'25
& $1\!\times\!3\!\times\!256^2$
& 1.44M & 2.29G
& 50.0 & 51.6 & \underline{99.8} & 0.2
& 57.9 & 58.3 & 59.5 & 56.2 \\

Effort~\cite{Yan_2025_Effort}
& ICML'25
& $1\!\times\!3\!\times\!224^2$
& 427.62M & 77.82G
& 47.6 & 45.8 & 86.0 & 9.3
& 70.3 & 74.3 & 65.3 & 75.3 \\

BSF~\cite{Zhou2026BSF}
& AAAI'26
& $1\!\times\!3\!\times\!224^2$
& 1.44M & 1.75G
& 50.0 & 45.9 & \textbf{99.9} & 0.1
& 56.1 & 61.3 & 36.5 & 75.8 \\

DFFreq~\cite{Yan_2026_DFFreq}
& TIFS'26
& $1\!\times\!3\!\times\!224^2$
& 1.44M & 1.77G
& 50.1 & 50.7 & 99.6 & 0.5
& 66.1 & 69.0 & 70.4 & 61.7 \\

LTD~\cite{Yang_2026_LTD}
& CVPR'26
& $1\!\times\!3\!\times\!224^2$
& 440.23M & 162.30G
& 50.0 & 49.4 & 96.3 & 3.7
& \underline{87.1} & 93.8 & 76.4 & \textbf{97.8} \\

IAPL~\cite{Li_2026_IAPL}
& CVPR'26
& $1\!\times\!3\!\times\!224^2$
& 343.75M & 167.53G
& 38.8 & 41.2 & 38.6 & \underline{39.1}
& 82.8 & 87.7 & 85.7 & 79.9 \\

PPM-CLIP~\cite{Wang_2026_PPMCLIP}
& CVPR'26
& $1\!\times\!3\!\times\!224^2$
& 453.41M & 694.38G
& -- & -- & -- & --
& 69.9 & 76.8 & 67.7 & 72.1 \\

\midrule

\rowcolor[RGB]{224,241,255}
\textbf{MAFDNet (Ours)}
& --
& $1\!\times\!3\!\times\!256^2$
& 1.04M & 1.39G
& -- & -- & -- & --
& \textbf{88.5} & \textbf{94.9} & \textbf{89.7} & 87.3 \\

\bottomrule
\end{tabularx}
\end{table*}

\begin{figure*}[t]
    \centering
    \includegraphics[width=0.92\textwidth]{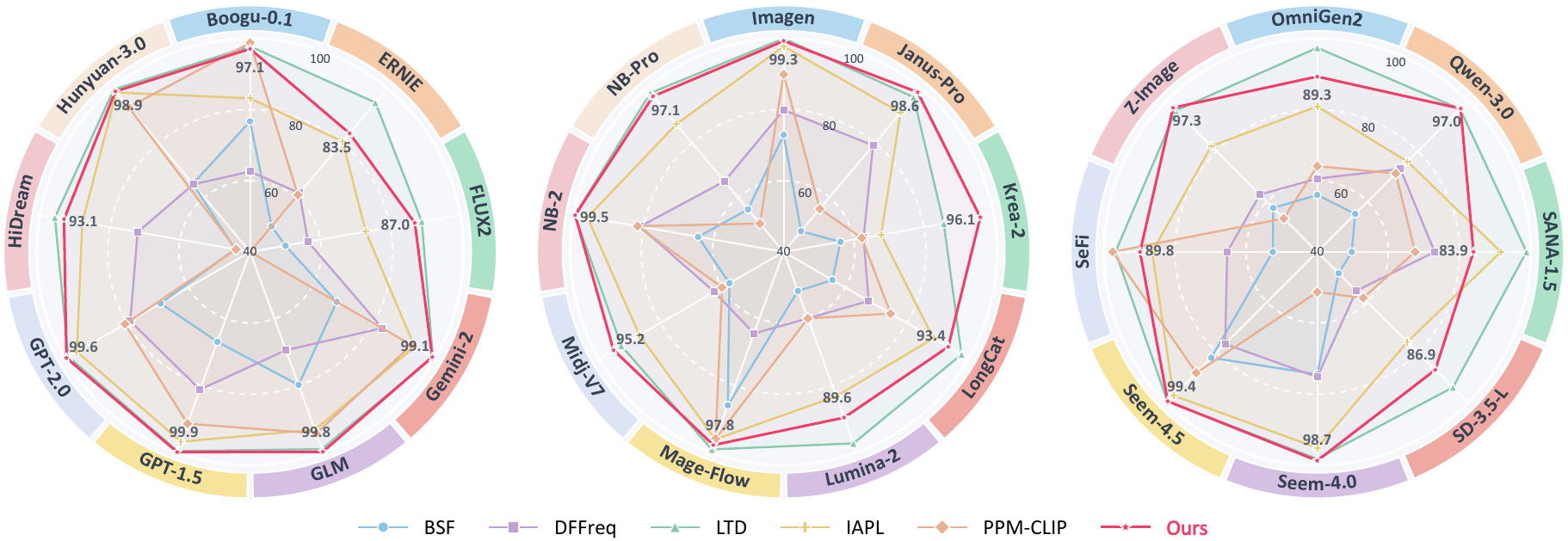}
    \caption{AP (\%) on the 26 generators in AIGI-26, grouped into three panels for readability. Numeric annotations show only MAFDNet results.}
    \label{fig:generator_ap}
\end{figure*}

\subsection{Experimental Setup}
\label{sec:exp_setup}

\noindent\textbf{Implementation.}
Experiments use one NVIDIA RTX A6000 GPU (48\,GB), CUDA 13.2, PyTorch 2.12.1, and Python 3.10.20.
We train MAFDNet for 150 epochs with batch size 128 using AdamW~\cite{Loshchilov_2019_AdamW}, $\beta=(0.9,0.999)$, initial learning rate $10^{-3}$, and weight decay $10^{-4}$.
A step schedule multiplies the learning rate by $0.9$, and gradients are clipped to norm $1.0$.
All images are uniformly re-encoded as JPEG with metadata removed to reduce source-dependent preprocessing bias.
Following AIGIBench~\cite{Li_2025_AIGIBench}, MAFDNet uses random crops during training and center crops during testing, both $256\times256$.
We use the final epoch-150 checkpoint and classify an image as generated when its sigmoid score exceeds $0.5$.
The default width is $c=40$.

\noindent\textbf{Metrics and baselines.}
We report average precision (AP), accuracy (Acc)~\cite{Tan_2024_FreqNet,Li_2025_SAFE,Yan_2025_Effort}, and real/generated class accuracies (R-Acc/F-Acc)~\cite{Yan_2025_AIDE,Wang_2026_Fleet}.
Table~\ref{tab1} compares 16 baselines using available official checkpoints and AIGI-26 retraining (Sec.~\ref{sec:data_protocol}), following AIGIBench configurations~\cite{Li_2025_AIGIBench}. Reproduction details are in the supplementary material.

\begin{table}[t]
\caption{Cross-dataset detection performance using model weights trained on AIGI-26.}
\label{tab:cross_dataset}
\centering
\fontsize{8pt}{9.2pt}\selectfont
\setlength{\tabcolsep}{3pt}
\begin{tabularx}{\columnwidth}{@{}l*{4}{>{\centering\arraybackslash}X}@{}}
\toprule
\multicolumn{5}{c}{\textbf{DailyBench}} \\
\midrule
Method & Acc$\uparrow$ & AP$\uparrow$ & R-Acc$\uparrow$ & F-Acc$\uparrow$ \\
\midrule
ResNet-50~\cite{He_2016_ResNet} & 57.7 & 61.0 & 41.0 & 74.4 \\
CNND~\cite{Wang_2020_CNNDetection} & \underline{66.1} & \textbf{76.6} & 48.8 & 83.5 \\
GramNet~\cite{Liu_2020_GramNet} & 57.3 & 62.2 & 37.7 & 76.8 \\
LGrad~\cite{Tan_2023_LGrad} & 53.8 & 53.4 & 47.3 & 60.2 \\
UnivFD~\cite{Ojha_2023_UniversalFakeDetect} & 52.9 & 58.4 & 14.2 & \underline{91.6} \\
NPR~\cite{Tan_2024_NPR} & 63.1 & 67.5 & \textbf{77.1} & 49.1 \\
LaDeDa~\cite{Cavia_2024_LaDeDa} & 51.9 & 53.6 & 67.0 & 36.9 \\
FreqNet~\cite{Tan_2024_FreqNet} & 49.7 & 49.5 & 67.4 & 31.9 \\
AIDE~\cite{Yan_2025_AIDE} & 52.3 & 59.8 & 20.7 & 83.9 \\
SAFE~\cite{Li_2025_SAFE} & 51.1 & 53.9 & 37.8 & 64.5 \\
Effort~\cite{Yan_2025_Effort} & 53.3 & 55.2 & 43.4 & 63.3 \\
BSF~\cite{Zhou2026BSF} & 49.6 & 52.4 & 37.4 & 61.8 \\
DFFreq~\cite{Yan_2026_DFFreq} & 60.4 & 62.2 & 60.8 & 59.9 \\
LTD~\cite{Yang_2026_LTD} & 51.9 & 59.8 & 9.9 & \textbf{94.0} \\
IAPL~\cite{Li_2026_IAPL} & 59.6 & 68.1 & 28.0 & 91.2 \\
PPM-CLIP~\cite{Wang_2026_PPMCLIP} & 53.1 & 55.5 & 15.0 & 91.3 \\
\midrule
\rowcolor[RGB]{224,241,255}
\textbf{MAFDNet (Ours)} & \textbf{69.8} & \underline{73.3} & \underline{74.3} & 65.3 \\
\midrule
\multicolumn{5}{c}{\textbf{Treasure-37}} \\
\midrule
Method & Acc$\uparrow$ & AP$\uparrow$ & R-Acc$\uparrow$ & F-Acc$\uparrow$ \\
\midrule
ResNet-50~\cite{He_2016_ResNet} & 59.9 & 66.9 & 40.7 & 79.1 \\
CNND~\cite{Wang_2020_CNNDetection} & \underline{68.0} & 79.4 & 56.0 & 80.0 \\
GramNet~\cite{Liu_2020_GramNet} & 58.2 & 64.7 & 36.1 & 80.3 \\
LGrad~\cite{Tan_2023_LGrad} & 55.4 & 55.8 & 48.5 & 62.3 \\
UnivFD~\cite{Ojha_2023_UniversalFakeDetect} & 62.0 & 77.2 & 32.5 & 91.4 \\
NPR~\cite{Tan_2024_NPR} & 67.9 & 74.6 & 56.4 & 79.3 \\
LaDeDa~\cite{Cavia_2024_LaDeDa} & 67.0 & 74.2 & \textbf{80.7} & 53.3 \\
FreqNet~\cite{Tan_2024_FreqNet} & 55.7 & 56.4 & \underline{73.4} & 37.9 \\
AIDE~\cite{Yan_2025_AIDE} & 64.1 & 76.4 & 35.7 & 92.5 \\
SAFE~\cite{Li_2025_SAFE} & 56.5 & 55.8 & 41.5 & 71.4 \\
Effort~\cite{Yan_2025_Effort} & 58.4 & 62.0 & 49.8 & 67.0 \\
BSF~\cite{Zhou2026BSF} & 62.3 & 66.4 & 36.3 & 88.3 \\
DFFreq~\cite{Yan_2026_DFFreq} & 62.4 & 62.3 & 55.1 & 69.6 \\
LTD~\cite{Yang_2026_LTD} & 66.3 & 80.1 & 36.3 & \textbf{96.3} \\
IAPL~\cite{Li_2026_IAPL} & 66.8 & 79.0 & 41.1 & 92.5 \\
PPM-CLIP~\cite{Wang_2026_PPMCLIP} & 67.4 & \textbf{85.4} & 41.5 & \underline{93.3} \\
\midrule
\rowcolor[RGB]{224,241,255}
\textbf{MAFDNet (Ours)} & \textbf{68.3} & \underline{80.4} & 52.6 & 84.0 \\
\bottomrule
\end{tabularx}
\end{table}

\begin{table*}[t]
\caption{Partial unseen-generator results on DailyBench and Treasure-37 using AIGI-26-trained models.}
\label{tab:unseen_generators}
\centering
\setlength{\tabcolsep}{2pt}
\fontsize{8pt}{9.8pt}\selectfont

\begin{tabularx}{\textwidth}{
@{}l*{4}{>{\centering\arraybackslash}X}
*{14}{>{\centering\arraybackslash}X}@{}}
\toprule

\multirow{3}{*}{Method}
& \multicolumn{4}{c}{\textbf{DailyBench}}
& \multicolumn{14}{c}{\textbf{Treasure-37}} \\

\cmidrule(lr){2-5}
\cmidrule(lr){6-19}

& \multicolumn{2}{c}{FLUX1}
& \multicolumn{2}{c}{Qwen-2512}
& \multicolumn{2}{c}{Kolors}
& \multicolumn{2}{c}{NextStep-1}
& \multicolumn{2}{c}{\shortstack{Playg-V2}}
& \multicolumn{2}{c}{Ovis-U1}
& \multicolumn{2}{c}{\shortstack{Wan2.2-F}}
& \multicolumn{2}{c}{BRIA-3.2}
& \multicolumn{2}{c}{Sora} \\

\cmidrule(lr){2-3}
\cmidrule(lr){4-5}
\cmidrule(lr){6-7}
\cmidrule(lr){8-9}
\cmidrule(lr){10-11}
\cmidrule(lr){12-13}
\cmidrule(lr){14-15}
\cmidrule(lr){16-17}
\cmidrule(lr){18-19}

& Acc & AP
& Acc & AP
& Acc & AP
& Acc & AP
& Acc & AP
& Acc & AP
& Acc & AP
& Acc & AP
& Acc & AP \\

\midrule

AIDE~\cite{Yan_2025_AIDE}
& 51.8 & 53.1
& 53.7 & 60.4
& 65.0 & 79.6
& 69.3 & 87.7
& 59.0 & 65.9
& 67.7 & 82.9
& 59.3 & 68.0
& 58.5 & 64.5
& 69.0 & \underline{88.2} \\

BSF~\cite{Zhou2026BSF}
& 44.7 & 45.1
& 46.1 & 47.1
& 63.0 & 69.4
& 66.1 & 56.2
& 65.6 & 63.1
& 62.0 & 65.1
& 64.9 & 68.5
& 61.8 & 68.2
& 62.5 & 69.6 \\

DFFreq~\cite{Yan_2026_DFFreq}
& \underline{59.5} & 61.3
& \underline{60.6} & 63.2
& 65.5 & 65.1
& 70.1 & 65.4
& 59.0 & 53.5
& 62.0 & 64.4
& 64.2 & 65.3
& 60.6 & 60.5
& \underline{70.3} & 76.7 \\

LTD~\cite{Yang_2026_LTD}
& 53.7 & \underline{71.4}
& 52.7 & 64.5
& \textbf{74.2} & 90.9
& \textbf{71.7} & \textbf{94.8}
& \textbf{72.8} & \underline{90.2}
& 66.5 & \textbf{92.8}
& \textbf{71.3} & \underline{87.5}
& 71.8 & 87.8
& \textbf{71.5} & 77.5 \\

IAPL~\cite{Li_2026_IAPL}
& 57.8 & 68.6
& 59.4 & \underline{72.4}
& 70.3 & \underline{92.5}
& 67.3 & 89.1
& 70.6 & 89.9
& \underline{69.2} & 90.2
& 67.7 & 72.4
& 69.3 & \underline{89.5}
& 65.2 & 68.9 \\

PPM-CLIP~\cite{Wang_2026_PPMCLIP}
& 53.0 & 56.4
& 54.1 & 59.1
& \underline{72.7} & \textbf{92.6}
& \underline{70.8} & 89.9
& 68.4 & \textbf{90.3}
& 67.5 & \underline{91.5}
& \underline{68.9} & 86.4
& \textbf{73.8} & \textbf{92.2}
& 65.2 & 87.1 \\

\midrule

\rowcolor[RGB]{224,241,255}
\textbf{MAFDNet (Ours)}
& \textbf{74.0} & \textbf{80.7}
& \textbf{81.6} & \textbf{90.7}
& 65.6 & 81.5
& \textbf{71.7} & \underline{94.2}
& \underline{72.5} & 76.8
& \textbf{70.3} & 85.2
& 68.6 & \textbf{88.2}
& \underline{71.9} & 82.6
& 70.1 & \textbf{92.6} \\

\bottomrule
\end{tabularx}
\end{table*}

\begin{table}[t]
\caption{Component ablations on AIGI-26. Each variant changes only the indicated component(s) of the full model.}
\label{tab:ablation_components}
\centering
\fontsize{8pt}{9.8pt}\selectfont
\setlength{\tabcolsep}{3pt}
\begin{tabularx}{\columnwidth}{@{}>{\raggedright\arraybackslash}X*{4}{>{\centering\arraybackslash}p{0.105\columnwidth}}@{}}
\toprule
Variant & Acc & AP & R-Acc & F-Acc \\
\midrule
\rowcolor{black!5}
\multicolumn{5}{c}{\itshape Complementary branches} \\
Spectral branch only & 70.6 & 76.9 & 60.4 & 80.8 \\
Spatial branch only & 80.9 & 88.3 & 78.8 & 83.1 \\
\midrule
\rowcolor{black!5}
\multicolumn{5}{c}{\itshape Forensic evidence} \\
w/o SRM filter & 83.3 & 90.4 & 84.3 & 82.2 \\
w/o Laplace filter & 81.2 & 88.6 & 77.7 & 84.7 \\
w/o Local residual & 82.1 & 89.7 & 82.4 & 81.7 \\
w/o Resampling residual & 81.3 & 89.4 & 78.8 & 83.7 \\
w/o Chroma difference & 83.0 & 90.5 & 82.9 & 83.1 \\
\midrule
\rowcolor{black!5}
\multicolumn{5}{c}{\itshape Spatial statistics and attention} \\
Mean only (w/o Std) & 87.8 & 94.6 & 88.1 & 87.5 \\
Std only (w/o Mean) & 87.3 & 94.6 & 87.3 & 87.2 \\
w/o LCA & 88.1 & 94.8 & 87.9 & 88.4 \\
\midrule
\rowcolor{black!5}
\multicolumn{5}{c}{\itshape Spectral statistics} \\
w/o Log compression & 87.1 & 94.5 & 88.1 & 86.0 \\
w/o Phase differences & 87.3 & 93.6 & 83.2 & \textbf{91.4} \\
w/o Radial masks & 84.8 & 92.8 & \textbf{89.7} & 80.0 \\
w/o Directional masks & 85.5 & 89.4 & 84.0 & 86.9 \\
\midrule
\rowcolor{lightblue}
\textbf{MAFDNet (Ours)} & \textbf{88.5} & \textbf{94.9} & \textbf{89.7} & 87.3 \\
\bottomrule
\end{tabularx}
\par\smallskip
\end{table}

\subsection{Comparison with Existing Detectors}
\label{sec:exp_comparison}

\noindent\textbf{Official pretrained model evaluation.}
Most of the 15 official pretrained models in Table~\ref{tab1} perform near chance (50\% Acc), with AP no higher than 59.2\%.
AIDE, SAFE, and BSF have at most 0.2\% F-Acc, missing almost all generated images.
Low AP also indicates weak score separation, so the degradation cannot be explained solely by the chosen decision threshold. This suggests limitations of earlier training data on modern generated images, without implying detector ineffectiveness.
Even the 144K ProGAN~\cite{Karras_2018_ProGAN} and SD-v1.4~\cite{Rombach_2022_LDM} images used by AIGIBench~\cite{Li_2025_AIGIBench} cannot ensure reliable detection of recent outputs, motivating our evaluation and retraining.

\noindent\textbf{Detection performance after retraining.}
Gains after retraining show that poor official-checkpoint results need not reflect architectural limits, supporting comparison under a shared contemporary data protocol.
MAFDNet leads in Acc/AP at 88.5\%/94.9\%, exceeding the strongest competing scores by 1.4/0.4 points.
Its 89.7\%/87.3\% R-Acc/F-Acc gives a 2.4-point class gap versus LTD's 21.4 points.
These gains reflect balanced class recognition and effective ranking, supporting the combination of local forensic and global spectral cues.

\noindent\textbf{Generator-wise performance.}
MAFDNet achieves 83.5--99.9\% AP across the 26 generators in AIGI-26, reaching at least 95\% on 17 (Fig.~\ref{fig:generator_ap}).
GPT-Image-1.5 and GLM-Image reach 99.9\% and 99.8\%, while ERNIE-Image (83.5\%) and SANA-1.5 (83.9\%) remain challenging.
LTD leads on OmniGen2 and SANA-1.5, indicating generator-dependent strengths. MAFDNet's strong AP across most generators supports broad effectiveness beyond a few easy sources.

\begin{figure}[t]
    \centering
    \includegraphics[width=0.9\linewidth]{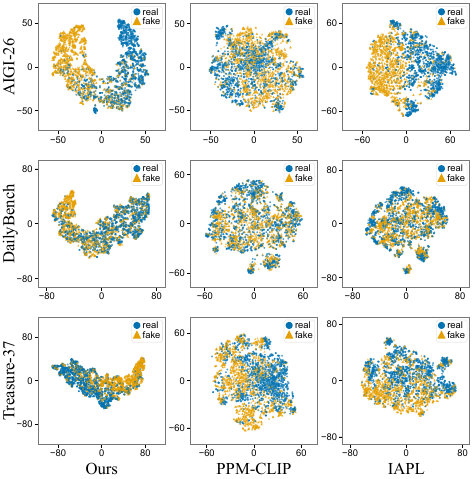}
    \caption{t-SNE feature visualizations of MAFDNet (Ours), PPM-CLIP, and IAPL on AIGI-26, DailyBench, and Treasure-37.}
    \label{fig:feature_tsne}
\end{figure}

\subsection{Cross-Dataset Generalization}
\label{sec:exp_generalization}

\noindent\textbf{Cross-dataset performance.}
Treasure-37 contains the 37 newly generated categories in Treasure~\cite{Wang_2026_Fleet}, excluding imports from GenImage~\cite{Zhu_2023_GenImage}, WildFake~\cite{Hong_2025_WildFake}, and MPBench~\cite{Lu_2023_MPBench}, among others. The generator list is in the supplementary material.
We test AIGI-26-trained weights on DailyBench~\cite{Jiang_2026_DailyBench} and Treasure-37.
MAFDNet achieves the best Acc (69.8\%/68.3\%) and second-best AP on both datasets (Table~\ref{tab:cross_dataset}), with a large class gap on Treasure-37.
Its Acc margins over the strongest competing results are 3.7 points on DailyBench and 0.3 points on Treasure-37, showing competitive accuracy across both external datasets. These results demonstrate effective cross-dataset generalization of the forensic cues learned on AIGI-26.

\noindent\textbf{Unseen-generator generalization.}
Table~\ref{tab:unseen_generators} presents partial unseen-generator results, with full results in the supplementary material.
MAFDNet leads in Acc/AP on DailyBench's FLUX1 and Qwen-2512 subsets, with Acc margins of 14.5/21.0 points over the strongest listed competitors.
On Treasure-37, it achieves the best Ovis-U1 Acc (70.3\%), joint-best NextStep-1 Acc, and best AP on Wan2.2-F and Sora.
These results demonstrate effective generalization beyond the training sources through transferable forensic cues.

\noindent\textbf{Feature visualization.}
The t-SNE projections~\cite{Maaten_2008_TSNE} in Fig.~\ref{fig:feature_tsne} show distinct class concentrations for MAFDNet and IAPL on AIGI-26, versus greater mixing for PPM-CLIP.
All three overlap more on DailyBench, although MAFDNet retains a generated-image-dominated region.
On Treasure-37, MAFDNet shows a concentrated generated-image region, and the alternatives also retain class-associated regions.
This structure is consistent with transferable cues, while increased mixing reflects ambiguity under domain shifts.

\subsection{Ablation Studies}
\label{sec:exp_ablation}

We ablate branches, encoder width, forensic cues, spatial statistics, attention, and spectral operations under the same AIGI-26 training protocol.

\noindent\textbf{Complementary branches.}
Fusion improves Acc/AP over the stronger spatial branch by 7.6/6.6 points and raises both class accuracies (Table~\ref{tab:ablation_components}).
Despite lower standalone accuracy, the spectral branch improves fusion, confirming that local and global evidence complement each other in recognizing both real and generated images.
Fusion benefits real-image recognition more strongly, narrowing the class gap.

\noindent\textbf{Encoder width.}
Width $c=40$ achieves the best Acc/AP. Width $c=32$ lowers Acc/AP by 0.4/0.5 points, whereas widths 48--96 increase model capacity without improving either metric (Table~\ref{tab:channel}).
This supports $c=40$, as greater capacity alone does not improve detection under this protocol.
The drop at $c=16$ suggests limited capacity.

\begin{table}[t]
\caption{Effect of base channel width $c$.}
\label{tab:channel}
\centering
\setlength{\tabcolsep}{3pt}
\fontsize{8pt}{9.8pt}\selectfont
\begin{tabularx}{\columnwidth}{@{}ccc*{4}{>{\centering\arraybackslash}X}@{}}
\toprule
Channel ($c$)
& Params
& FLOPs
& Acc
& AP
& R-Acc
& F-Acc \\
\midrule

16
& 0.179M & 0.431G
& 78.6 & 86.1 & 76.0 & 81.2 \\

32
& 0.672M & 0.997G
& 88.1 & 94.4 & 86.9 & \textbf{89.2} \\

\rowcolor[RGB]{224,241,255}
40
& 1.036M & 1.387G
& \textbf{88.5} & \textbf{94.9} & \textbf{89.7} & 87.3 \\

48
& 1.478M & 1.848G
& 87.7 & 94.8 & 87.0 & 88.3 \\

64
& 2.598M & 2.985G
& 84.4 & 91.7 & 84.4 & 84.5 \\

96
& 5.780M & 6.114G
& 84.7 & 92.5 & 89.2 & 80.2 \\

\bottomrule
\end{tabularx}
\end{table}

\noindent\textbf{Forensic evidence.}
Removing any non-RGB group lowers Acc by 5.2--7.3 points and AP by 4.4--6.3 points, with Laplace removal most detrimental (Table~\ref{tab:ablation_components}).
The consistent losses indicate that explicit forensic responses provide useful evidence beyond RGB features alone.
The remaining cues do not fully compensate for each removed response.

\noindent\textbf{Spatial statistics and attention.}
Joint mean/std pooling improves Acc by 0.7/1.2 points over mean-only/std-only pooling (Table~\ref{tab:ablation_components}).
LCA adds a modest 0.4/0.1-point Acc/AP gain, improving real-image recognition at a slight cost to generated-image recognition.
Together, these results support complementary response-strength and dispersion cues, with channel reweighting providing a smaller refinement.
Response dispersion thus adds information beyond average activation strength.

\noindent\textbf{Spectral statistics.}
All spectral ablations reduce Acc/AP (Table~\ref{tab:ablation_components}). Removing phase differences widens the class gap.
Radial-mask removal causes the largest Acc drop (3.7 points), while directional-mask removal causes the largest AP drop (5.5 points).
These losses support preserving frequency-scale and directional structure when aggregating spectral evidence.
Log compression improves both class accuracies, consistent with limiting magnitude dominance.

\section{Conclusion}
\label{sec:conclusion}

We introduce AIGI-26, a benchmark of 253,200 images from 26 recent generators, and MAFDNet, which combines spatial and spectral forensic evidence.
After retraining on AIGI-26, MAFDNet achieves 88.5\% Acc and 94.9\% AP with 1.04M parameters and 1.39 GFLOPs.
Cross-dataset evaluations demonstrate the relative transferability of the learned forensic representations. 
Ablation studies further confirm the contributions of fixed forensic responses, multiscale spatial statistics, and spectral cues to detection performance.
Future work explores real-world robustness and advances fine-grained forensic evidence attribution.


{
    \small
    \bibliographystyle{ieeenat_fullname}
    \bibliography{main}
}

\clearpage
\setcounter{page}{1}
\renewcommand{\thepage}{S\arabic{page}}
\setcounter{section}{0}
\setcounter{equation}{0}
\setcounter{table}{0}
\setcounter{figure}{0}
\renewcommand{\thesection}{\Alph{section}}
\renewcommand{\theequation}{S\arabic{equation}}
\renewcommand{\thetable}{S\arabic{table}}
\renewcommand{\thefigure}{S\arabic{figure}}
\renewcommand{\theHsection}{supp.\Alph{section}}
\renewcommand{\theHequation}{supp.\arabic{equation}}
\renewcommand{\theHtable}{supp.\arabic{table}}
\renewcommand{\theHfigure}{supp.\arabic{figure}}
\maketitlesupplementary
\raggedbottom
\definecolor{suppheader}{RGB}{242,245,248}
\definecolor{suppours}{RGB}{224,239,252}
\makeatletter
\setlength{\@fptop}{0pt}
\setlength{\@dblfptop}{0pt}
\setlength{\@fpsep}{12pt}
\setlength{\@dblfpsep}{12pt}
\makeatother
\setcounter{topnumber}{4}
\setcounter{dbltopnumber}{4}
\setcounter{totalnumber}{6}
\renewcommand{\topfraction}{0.95}
\renewcommand{\dbltopfraction}{0.95}
\renewcommand{\textfraction}{0.05}
\renewcommand{\floatpagefraction}{0.65}
\renewcommand{\dblfloatpagefraction}{0.65}
\setlength{\dbltextfloatsep}{12pt plus 2pt minus 2pt}
\setlength{\textfloatsep}{12pt plus 2pt minus 2pt}

\section*{Overview}
This supplementary material provides additional details on AIGI-26 and the evaluation of MAFDNet. It is organized as follows:
\begin{itemize}\setlength{\itemsep}{2pt}\setlength{\parskip}{0pt}
    \item \textbf{Sec.~\ref{sec:supp_data}} describes the generators, dataset composition, partitioning, and quality control.
    \item \textbf{Sec.~\ref{sec:supp_protocol}} lists the comparison detectors and external evaluation datasets.
    \item \textbf{Sec.~\ref{sec:supp_results}} reports detailed results across image sizes, generators, and datasets, followed by robustness to image degradations and feature visualizations.
\end{itemize}

\section{AIGI-26: Sources and Construction}
\label{sec:supp_data}
\subsection{Generator Inventory}
AIGI-26 combines images from 26 contemporary generators, including 16 open-source and ten closed-source models. Table~\ref{tab:supp_generators} lists their versions, acquisition routes, and model resources. We deploy 15 models locally using the weights linked in the table and access six models through APIs. HunyuanImage-3.0 is open-source but is accessed through an API in our collection.
We additionally reuse five subsets from T2I-CoReBench~\cite{Li2026T2ICoReBench}: Seedream-4.5, GPT-Image-1.5, Imagen 4 Ultra, Gemini 2.0 Flash, and Nano Banana 2. Each contributes 4,320 fake images. These existing outputs are used for authenticity detection rather than their original compositional and reasoning evaluation. We use the 2025 release of SD 3.5-L and the 2026 release of Z-Image-Turbo.

\begin{table*}[!t]
\centering
\caption{Generators used in AIGI-26, grouped by acquisition route. Local-model links point to the Hugging Face weights used for generation. Dates identify the versions used in our collection, with months shown where recorded.}
\label{tab:supp_generators}
\small\setlength{\tabcolsep}{4pt}
\renewcommand{\arraystretch}{1.12}
\begin{tabularx}{\textwidth}{@{}r p{0.27\textwidth} >{\centering\arraybackslash}p{0.11\textwidth} >{\centering\arraybackslash}p{0.15\textwidth} X@{}}
\toprule
\rowcolor{suppheader}
\textbf{ID} & \textbf{Generator} & \textbf{Params.} & \textbf{Version date} & \textbf{Weights / model resource} \\
\midrule
\rowcolor{suppheader}
\multicolumn{5}{l}{\textbf{Local deployment (15 models, 5,000 fake images per model)}}\\
1 & SANA-1.5 & 4.8B & 2025 & \href{https://huggingface.co/Efficient-Large-Model/SANA1.5_4.8B_1024px_diffusers}{Efficient-Large-Model / SANA1.5-4.8B} \\
2 & Janus-Pro & 7B & 2025 & \href{https://huggingface.co/deepseek-ai/Janus-Pro-7B}{deepseek-ai / Janus-Pro-7B} \\
3 & Lumina-Image-2.0 & 2.6B & 2025 & \href{https://huggingface.co/Alpha-VLLM/Lumina-Image-2.0}{Alpha-VLLM / Lumina-Image-2.0} \\
4 & SD 3.5-L & 8.1B & 2025 & \href{https://huggingface.co/stabilityai/stable-diffusion-3.5-large}{stabilityai / stable-diffusion-3.5-large} \\
5 & OmniGen2 & 7B & 2025 & \href{https://huggingface.co/OmniGen2/OmniGen2}{OmniGen2 / OmniGen2} \\
6 & LongCat-Image & 13B & 2025 & \href{https://huggingface.co/meituan-longcat/LongCat-Image}{meituan-longcat / LongCat-Image} \\
7 & Z-Image-Turbo & 6B & 2026 & \href{https://huggingface.co/Tongyi-MAI/Z-Image-Turbo}{Tongyi-MAI / Z-Image-Turbo} \\
8 & Mage-Flow-Turbo & 4B & 2026 & \href{https://huggingface.co/microsoft/Mage-Flow-Turbo}{microsoft / Mage-Flow-Turbo} \\
9 & GLM-Image & 16B & 2026 & \href{https://huggingface.co/zai-org/GLM-Image}{zai-org / GLM-Image} \\
10 & Boogu-Image-0.1-Turbo & 10B & 2026 & \href{https://huggingface.co/Boogu/Boogu-Image-0.1-Turbo}{Boogu / Boogu-Image-0.1-Turbo} \\
11 & FLUX.2-klein & 4B & 2026.01 & \href{https://huggingface.co/black-forest-labs/FLUX.2-klein-4B}{black-forest-labs / FLUX.2-klein-4B} \\
12 & ERNIE-Image & 8B & 2026.05 & \href{https://huggingface.co/baidu/ERNIE-Image}{baidu / ERNIE-Image} \\
13 & HiDream-O1-Image & 8B & 2026.05 & \href{https://huggingface.co/HiDream-ai/HiDream-O1-Image}{HiDream-ai / HiDream-O1-Image} \\
14 & Krea-2-Turbo & 12B & 2026.06 & \href{https://huggingface.co/krea/Krea-2-Turbo}{krea / Krea-2-Turbo} \\
15 & SeFi-Image-Turbo & 5B & 2026.06 & \href{https://huggingface.co/SeFi-Image/SeFi-Image-5B-turbo}{SeFi-Image / SeFi-Image-5B-turbo} \\
\midrule
\rowcolor{suppheader}
\multicolumn{5}{l}{\textbf{API generation (6 models, 5,000 fake images per model)}}\\
16 & GPT-Image-2 & -- & 2026 & \href{https://developers.openai.com/api/docs/models/gpt-image-2}{OpenAI / GPT-Image-2} \\
17 & Qwen-Image-3.0 & -- & 2026 & \href{https://www.alibabacloud.com/help/en/model-studio/image-model/}{Alibaba / Qwen image API} \\
18 & Seedream-4.0 & -- & 2025 & \href{https://seed.bytedance.com/en/seedream4_0}{ByteDance / Seedream-4.0} \\
19 & Nano Banana Pro & -- & 2025 & \href{https://ai.google.dev/gemini-api/docs/models}{Google / Gemini image API} \\
20 & Midjourney V7 & -- & 2025 & \href{https://docs.midjourney.com/hc/en-us/articles/32199405667853-Version}{Midjourney / V7} \\
21 & HunyuanImage-3.0 & -- & 2025 & \href{https://huggingface.co/tencent/HunyuanImage-3.0}{Tencent / HunyuanImage-3.0} \\
\midrule
\rowcolor{suppheader}
\multicolumn{5}{l}{\textbf{Reused from T2I-CoReBench~\cite{Li2026T2ICoReBench} (5 models, 4,320 fake images per model)}}\\
22 & Seedream-4.5 & -- & 2025 & \href{https://seed.bytedance.com/en/seedream4_5}{ByteDance / Seedream-4.5} \\
23 & GPT-Image-1.5 & -- & 2025 & \href{https://developers.openai.com/api/reference/resources/images/methods/generate}{OpenAI / GPT-Image-1.5} \\
24 & Imagen 4 Ultra & -- & 2025 & \href{https://ai.google.dev/gemini-api/docs/models/imagen}{Google / Imagen 4 Ultra} \\
25 & Gemini 2.0 Flash & -- & 2025 & \href{https://ai.google.dev/gemini-api/docs/models}{Google / Gemini image API} \\
26 & Nano Banana 2 & -- & 2026 & \href{https://ai.google.dev/gemini-api/docs/models}{Google / Gemini image API} \\
\bottomrule
\end{tabularx}
\end{table*}

\subsection{Sample Counts and Partitioning}
AIGI-26 contains 253,200 images, equally divided into 126,600 fake and 126,600 real images. The 15 local models and six API models each contribute 5,000 fake images, while the five reused subsets each contribute 4,320. We retrieve an equal number of real images from LAION-Aesthetics V2~\cite{LAION_Aesthetics_V2}, matching topics and scenes to each fake subset. Table~\ref{tab:supp_acquisition} summarizes this composition and the final partition sizes.

\begin{table}[!t]
\centering\small
\caption{AIGI-26 sample composition and split. Real images are retrieved to match the fake subsets in each acquisition group.}
\label{tab:supp_acquisition}
\setlength{\tabcolsep}{3pt}
\renewcommand{\arraystretch}{1.12}
\begin{tabularx}{\columnwidth}{@{}Xrrr@{}}
\toprule
\rowcolor{suppheader}
\textbf{Source / split} & \textbf{Fake} & \textbf{Real} & \textbf{Total} \\
\midrule
Local (15 models) & 75,000 & 75,000 & 150,000 \\
API (6 models) & 30,000 & 30,000 & 60,000 \\
Reuse (5 models) & 21,600 & 21,600 & 43,200 \\
\midrule
Training (5\%) & 6,330 & 6,330 & 12,660 \\
Testing (95\%) & 120,270 & 120,270 & 240,540 \\
\midrule
\rowcolor{suppheader}
\textbf{All images} & \textbf{126,600} & \textbf{126,600} & \textbf{253,200} \\
\bottomrule
\end{tabularx}
\end{table}

The class-balanced training and test partitions contain 12,660 and 240,540 images, respectively. Fake images sharing a prompt across generators remain in the same partition. Real images are grouped by matched topics and scenes. Duplicate image IDs are removed before partitioning. Group-level sampling is stratified by generator and authenticity class, reducing repeated-content leakage while preserving comparable class proportions.

\begin{figure}[!t]
\centering
\includegraphics[width=0.9\columnwidth]{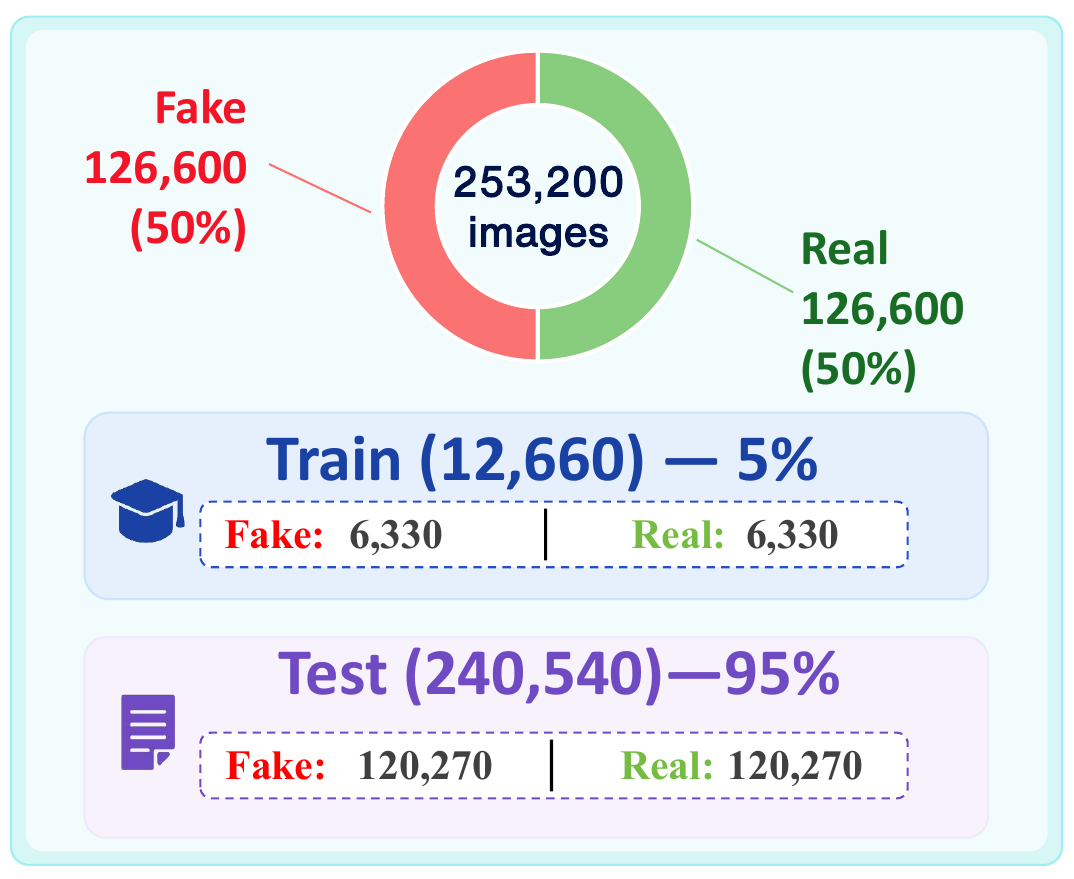}
\caption{AIGI-26 class composition and partition sizes. Both the 5\% training partition and the 95\% test partition contain equal numbers of real and fake images.}
\label{fig:supp_dataset}
\end{figure}

Figure~\ref{fig:supp_dataset} visualizes the equal real/fake composition in both partitions. The training subset supports learning from contemporary generators, while the larger test subset covers the collection's varied generation sources. Table~\ref{tab:supp_resolutions} reports the eight image-size settings, their sample counts, and representative uses. Square, portrait, and landscape outputs provide varied image layouts within the same detection task.

\begin{table*}[!t]
\centering\small
\caption{Image-size settings, sample counts, and representative application scenarios in AIGI-26. Format labels denote nominal aspect ratios. Counts include both authenticity classes.}
\label{tab:supp_resolutions}
\setlength{\tabcolsep}{4pt}
\renewcommand{\arraystretch}{1.12}
\begin{tabularx}{\textwidth}{@{}l >{\centering\arraybackslash}p{0.22\textwidth} >{\raggedleft\arraybackslash}p{0.09\textwidth} >{\raggedleft\arraybackslash}p{0.10\textwidth} >{\raggedleft\arraybackslash}p{0.10\textwidth} X@{}}
\toprule
\rowcolor{suppheader}
\textbf{Format} & \textbf{Resolution ($W\times H$)} & \textbf{Train} & \textbf{Test} & \textbf{Total} & \textbf{Representative uses} \\
\midrule
1:1 & $1024\times1024$ & 1,791 & 35,547 & 37,338 & General imagery, avatars, social media \\
2:3 & $832\times1216$ & 1,579 & 29,288 & 30,867 & Vertical portraits, magazine covers \\
3:2 & $1216\times832$ & 1,526 & 29,342 & 30,868 & Landscape photography, cinematic stills \\
3:4 & $896\times1152$ & 1,511 & 29,469 & 30,980 & Portrait photography, poster design \\
4:3 & $1152\times896$ & 1,546 & 29,318 & 30,864 & Photography displays, blog illustrations \\
9:16 & $768\times1344$ & 1,537 & 29,311 & 30,848 & Mobile wallpapers, short-form video assets \\
16:9 & $1344\times768$ & 1,522 & 29,220 & 30,742 & Desktop wallpapers, widescreen video covers \\
21:9 & $1536\times640$ & 1,648 & 29,045 & 30,693 & Cinematic panoramas, wide banners \\
\midrule
\rowcolor{suppheader}
\multicolumn{2}{l}{\textbf{Total}} & \textbf{12,660} & \textbf{240,540} & \textbf{253,200} & \\
\bottomrule
\end{tabularx}
\end{table*}

\subsection{Prompts, Real Images, and Quality Control}
\noindent\textbf{Prompt construction and screening.}
ChatGPT and Gemini provide 3,000 and 2,000 seed prompts, respectively, expanded through subject recombination and sentence transformations. We encode prompts with \href{https://huggingface.co/Qwen/Qwen3-Embedding-0.6B}{Qwen3-Embedding-0.6B}~\cite{Qwen3Embedding06B_Model} and identify semantic duplicates at a cosine-similarity threshold of 0.7. ChatGPT rewrites duplicate or unreasonable descriptions for rechecking before prompts are randomly grouped and assigned to generators.

\noindent\textbf{Matching-text preparation and retrieval.}
For the 21 directly accessed generators (15 local and six API models), the original generation prompts are matched against the \texttt{text} field of LAION-Aesthetics V2 candidate images. For the five reused T2I-CoReBench subsets, Qwen3-VL-4B-Instruct~\cite{Qwen3VL4B_Model} describes each fake image, and the description is matched using the same rule.

\noindent\textbf{Content-matching criterion.}
Agreement on two or three topic, keyword, or scene cues establishes a candidate match. We compare subjects and scene descriptions in the source text with the candidate's \texttt{text}, reducing content differences between real and fake images. Matching retains variation in viewpoints, backgrounds, and compositions. Candidates then undergo ID checks and image screening before inclusion.

\noindent\textbf{Image-ID checks.}
Image names serve as IDs. Candidate IDs are checked against accepted real images to prevent reuse across queries and generator subsets. Repeated IDs are rejected, and another eligible candidate is used. Both classes undergo ID checks before partitioning, followed by a check for overlap between training and test IDs.

\noindent\textbf{Image screening.}
\href{https://huggingface.co/Qwen/Qwen3-VL-4B-Instruct}{Qwen3-VL-4B-Instruct}~\cite{Qwen3VL4B_Model} screens both real and fake images for content completeness and visual quality. Blank or severely incomplete images, quality failures that prevent inspection, and retrieval mismatches receive human review. Clearly invalid images are excluded, while generation artifacts are retained as detection evidence. Screening assesses the image itself, independently of the text-matching criterion.

\noindent\textbf{Content-aware cropping and size allocation.}
Fake images are generated at the predefined aspect ratios and sizes. Real images are assigned to the same eight settings through balanced allocation and reproducible shuffling, and undergo content-aware cropping. Grayscale gradients and local contrast define a reduced-resolution importance map. An integral-image search selects a salient crop with a weak center preference, followed by Lanczos resizing. For the widest format, a sharp central crop is supplemented with reflected side regions, Gaussian smoothing, and feathered transitions. Each real input contributes one output, preserving salient subjects while balancing real and fake image sizes.

\noindent\textbf{JPEG standardization.}
We correct EXIF orientation, composite transparent pixels onto white, and copy pixels into a new RGB image, removing inherited EXIF, ICC, XMP, text comments, prompts, and software or generator metadata. All images use non-progressive JPEG encoding at quality 100, with 4:2:0 chroma subsampling and optimized encoding, reducing class-dependent format and metadata differences.

\section{Comparison Methods and Evaluation Protocol}
\label{sec:supp_protocol}
\subsection{Detector Inventory}
We compare MAFDNet with the 16 detectors in Table~\ref{tab:supp_detectors}. These methods cover convolutional and texture representations, local pixel and gradient evidence, frequency analysis, pretrained visual features, and prompt-based adaptation. Their complementary designs provide a broad comparison of the evidence used for AIGI detection. ResNet-50 uses the truncated architecture specified in the main paper. The 13 methods from ResNet-50 through DFFreq use the detector implementations provided by AIGIBench~\cite{Li_2025_AIGIBench}. LTD, IAPL, and PPM-CLIP use their respective official repositories.

We use available official checkpoints for direct evaluation and retrain the comparison methods on the AIGI-26 training partition. Retraining follows the released training configurations of the linked implementations without additional hyperparameter tuning. The two settings separate the applicability of existing trained models from detector performance after training on contemporary images. We report Acc, AP, R-Acc, and F-Acc under the evaluation protocol of the main paper.

\begin{table*}[!t]
\centering
\caption{Comparison detectors and implementation sources. ResNet-50 through DFFreq use the AIGIBench repository, while LTD, IAPL, and PPM-CLIP use their official repositories.}
\label{tab:supp_detectors}
\small\setlength{\tabcolsep}{4pt}
\renewcommand{\arraystretch}{1.06}
\begin{tabularx}{\textwidth}{@{}l >{\raggedright\arraybackslash}X >{\raggedright\arraybackslash}p{0.40\textwidth}@{}}
\toprule
\rowcolor{suppheader}
\textbf{Method} & \textbf{Principal mechanism} & \textbf{Code source} \\
\midrule
ResNet-50~\cite{He_2016_ResNet} & Truncated two-stage convolutional baseline & \href{https://github.com/HorizonTEL/AIGIBench/tree/main/detector_codes/Resnet50-main}{AIGIBench / ResNet-50} \\
CNND~\cite{Wang_2020_CNNDetection} & CNN-based artifact detection & \href{https://github.com/HorizonTEL/AIGIBench/tree/main/detector_codes/CNNDetection-master}{AIGIBench / CNND} \\
GramNet~\cite{Liu_2020_GramNet} & Global texture representations & \href{https://github.com/HorizonTEL/AIGIBench/tree/main/detector_codes/Gram-Net-main}{AIGIBench / GramNet} \\
LGrad~\cite{Tan_2023_LGrad} & Gradient-based artifact representations & \href{https://github.com/HorizonTEL/AIGIBench/tree/main/detector_codes/LGrad-master}{AIGIBench / LGrad} \\
UnivFD~\cite{Ojha_2023_UniversalFakeDetect} & Pretrained CLIP feature classification & \href{https://github.com/HorizonTEL/AIGIBench/tree/main/detector_codes/CLIPDetection-main}{AIGIBench / UnivFD} \\
NPR~\cite{Tan_2024_NPR} & Neighboring-pixel relationships & \href{https://github.com/HorizonTEL/AIGIBench/tree/main/detector_codes/NPR-DeepfakeDetection-main}{AIGIBench / NPR} \\
LaDeDa~\cite{Cavia_2024_LaDeDa} & Local evidence for efficient detection & \href{https://github.com/HorizonTEL/AIGIBench/tree/main/detector_codes/RealTime-DeepfakeDetection-in-the-RealWorld-main}{AIGIBench / LaDeDa} \\
FreqNet~\cite{Tan_2024_FreqNet} & Frequency-aware feature learning & \href{https://github.com/HorizonTEL/AIGIBench/tree/main/detector_codes/FreqNet-DeepfakeDetection-main}{AIGIBench / FreqNet} \\
AIDE~\cite{Yan_2025_AIDE} & Complementary semantic and forensic views & \href{https://github.com/HorizonTEL/AIGIBench/tree/main/detector_codes/AIDE-main}{AIGIBench / AIDE} \\
SAFE~\cite{Li_2025_SAFE} & Image transformations for transferable cues & \href{https://github.com/HorizonTEL/AIGIBench/tree/main/detector_codes/SAFE-main}{AIGIBench / SAFE} \\
Effort~\cite{Yan_2025_Effort} & Orthogonal subspace adaptation & \href{https://github.com/HorizonTEL/AIGIBench/tree/main/detector_codes/Effort-AIGI-Detection}{AIGIBench / Effort} \\
BSF~\cite{Zhou2026BSF} & Pixel-level mapping to reduce semantic bias & \href{https://github.com/HorizonTEL/AIGIBench/tree/main/detector_codes/BSF-main}{AIGIBench / BSF} \\
DFFreq~\cite{Yan_2026_DFFreq} & Dual frequency branches and window attention & \href{https://github.com/HorizonTEL/AIGIBench/tree/main/detector_codes/DFFreq-main}{AIGIBench / DFFreq} \\
LTD~\cite{Yang_2026_LTD} & Inter-layer latent transition discrepancies & \href{https://github.com/yywencs/LTD}{yywencs/LTD} \\
IAPL~\cite{Li_2026_IAPL} & Image-adaptive prompt learning & \href{https://github.com/liyih/IAPL}{liyih/IAPL} \\
PPM-CLIP~\cite{Wang_2026_PPMCLIP} & Probabilistic prompt modeling with CLIP & \href{https://github.com/bandaidssssss/PPM_CLIP}{bandaidssssss/PPM\_CLIP} \\
\bottomrule
\end{tabularx}
\end{table*}

\subsection{External Test Sets and Unseen Generators}
We directly evaluate AIGI-26-trained models on DailyBench~\cite{Jiang_2026_DailyBench} and Treasure-37~\cite{Wang_2026_Fleet}. Table~\ref{tab:supp_external_generators} lists the seven DailyBench text-to-image generators and the 37 Treasure generator categories used in these evaluations. DailyBench supplies contemporary, high-quality fake images. Treasure-37 comprises the 37 populated subsets in the \href{https://huggingface.co/datasets/ThreeLiu/Treasure\#included-fake-subsets}{\textcolor{black}{Treasure release}}, excluding imported collections and placeholder categories. The two datasets therefore extend evaluation to independently collected images and a broader range of generation sources.

\begin{table*}[!t]
\centering\small
\caption{Generator inventories for DailyBench and Treasure-37. IDs are local to each dataset. Treasure-37 follows the 37 self-generated subsets included in the public release.}
\label{tab:supp_external_generators}
\setlength{\tabcolsep}{5pt}
\renewcommand{\arraystretch}{1.06}
\begin{tabularx}{\textwidth}{@{}r X r X r X@{}}
\toprule
\rowcolor{suppheader}
\multicolumn{6}{c}{\textbf{DailyBench}~\cite{Jiang_2026_DailyBench}}\\
\midrule
\textbf{ID} & \textbf{Generator} & \textbf{ID} & \textbf{Generator} & \textbf{ID} & \textbf{Generator} \\
\midrule
1 & SD 3.5 Large & 4 & Qwen-Image-2512 & 7 & Nano Banana 2 \\
2 & FLUX.1 & 5 & Z-Image & & \\
3 & FLUX.2 & 6 & GPT-Image-2 & & \\
\midrule
\rowcolor{suppheader}
\multicolumn{6}{c}{\textbf{Treasure-37}~\cite{Wang_2026_Fleet}}\\
\midrule
\textbf{ID} & \textbf{Generator} & \textbf{ID} & \textbf{Generator} & \textbf{ID} & \textbf{Generator} \\
\midrule
1 & Playground V2 & 14 & BAGEL & 27 & Doubao Seedream 4.0 \\
2 & Playground V2.5 & 15 & BRIA 3.2 & 28 & Doubao Seedream 3.0 \\
3 & Hunyuan-DiT & 16 & OmniGen2 & 29 & HunyuanImage 3.0 \\
4 & LlamaGen & 17 & Show-o2 & 30 & FLUX.2 \\
5 & SD3-medium & 18 & Ovis-U1 & 31 & Wan2.2-T2I-Flash \\
6 & Show-o & 19 & NextStep-1 & 32 & Wan2.5-T2I-Preview \\
7 & OmniGen & 20 & Z-Image-Turbo & 33 & CogView4 \\
8 & CogView3plus & 21 & LongCat-Image & 34 & Sora-image \\
9 & Infinity-2B & 22 & Kolors & 35 & GPT-image-1.5 \\
10 & Janus-Pro-7B & 23 & Qwen-Image & 36 & Midjourney v6.1 \\
11 & SANA v1.5 & 24 & Imagen 4 & 37 & Midjourney v7 \\
12 & LUMINA-Image 2.0 & 25 & Nano Banana &  &  \\
13 & HiDream-I1-Dev & 26 & Nano Banana Pro &  &  \\
\bottomrule
\end{tabularx}
\end{table*}

\begin{table*}[!t]
\centering
\caption{Acc and AP (\%) across the eight image-size groups of the AIGI-26 test set using AIGI-26-trained models. \textbf{Bold} and \underline{underlined} values mark the best and second-best distinct scores in each metric column, including ties.}
\label{tab:supp_scale_results}
\fontsize{8.5}{10.5}\selectfont
\setlength{\tabcolsep}{1.5pt}
\renewcommand{\arraystretch}{1.18}
\begin{tabularx}{\textwidth}{@{}p{72pt}*{18}{>{\centering\arraybackslash}X}@{}}
\toprule
\rowcolor{suppheader}
\textbf{Method} & \multicolumn{2}{c}{\shortstack{1:1\\$1024\times1024$}} & \multicolumn{2}{c}{\shortstack{2:3\\$832\times1216$}} & \multicolumn{2}{c}{\shortstack{3:2\\$1216\times832$}} & \multicolumn{2}{c}{\shortstack{3:4\\$896\times1152$}} & \multicolumn{2}{c}{\shortstack{4:3\\$1152\times896$}} & \multicolumn{2}{c}{\shortstack{9:16\\$768\times1344$}} & \multicolumn{2}{c}{\shortstack{16:9\\$1344\times768$}} & \multicolumn{2}{c}{\shortstack{21:9\\$1536\times640$}} & \multicolumn{2}{c}{\textbf{Mean}} \\
\cmidrule(lr){2-3}\cmidrule(lr){4-5}\cmidrule(lr){6-7}\cmidrule(lr){8-9}\cmidrule(lr){10-11}\cmidrule(lr){12-13}\cmidrule(lr){14-15}\cmidrule(lr){16-17}\cmidrule(lr){18-19}
 & Acc & AP & Acc & AP & Acc & AP & Acc & AP & Acc & AP & Acc & AP & Acc & AP & Acc & AP & Acc & AP \\
\midrule
ResNet-50~\cite{He_2016_ResNet} & 69.3 & 69.3 & 71.0 & 77.0 & 71.1 & 77.5 & 71.9 & 78.8 & 72.2 & 78.3 & 70.6 & 76.9 & 70.5 & 76.3 & 69.7 & 74.6 & 70.8 & 76.1 \\
CNND~\cite{Wang_2020_CNNDetection} & 84.6 & 87.9 & 81.8 & 89.7 & 85.8 & \textbf{95.7} & 84.0 & 87.8 & 83.6 & 87.4 & 87.0 & 93.3 & 87.3 & 92.5 & 79.8 & 90.6 & 84.2 & 90.6 \\
GramNet~\cite{Liu_2020_GramNet} & 66.2 & 66.8 & 68.0 & 74.1 & 67.7 & 73.1 & 69.4 & 75.6 & 68.8 & 74.1 & 67.6 & 73.2 & 67.2 & 72.5 & 66.3 & 71.9 & 67.7 & 72.7 \\
LGrad~\cite{Tan_2023_LGrad} & 56.7 & 50.5 & 55.9 & 58.6 & 54.4 & 57.1 & 55.8 & 59.0 & 55.8 & 59.0 & 54.9 & 57.4 & 54.3 & 56.8 & 55.9 & 58.8 & 55.5 & 57.1 \\
UnivFD~\cite{Ojha_2023_UniversalFakeDetect} & 79.8 & 85.2 & 79.2 & 88.2 & 78.9 & 87.6 & 79.5 & 88.6 & 79.0 & 88.0 & 79.6 & 88.7 & 78.7 & 87.2 & 71.2 & 81.5 & 78.3 & 86.9 \\
NPR~\cite{Tan_2024_NPR} & 68.9 & 72.1 & 70.1 & 81.3 & 69.9 & 82.2 & 70.6 & 82.1 & 70.4 & 82.1 & 69.6 & 81.6 & 70.2 & 82.5 & 67.4 & 76.3 & 69.6 & 80.0 \\
LaDeDa~\cite{Cavia_2024_LaDeDa} & 79.0 & 83.7 & 86.5 & 93.9 & 86.2 & 93.8 & 85.9 & 93.5 & 85.9 & 93.7 & 86.7 & 94.5 & 87.2 & 94.5 & 75.4 & 85.0 & 84.1 & 91.6 \\
FreqNet~\cite{Tan_2024_FreqNet} & 64.8 & 62.7 & 68.0 & 76.1 & 67.8 & 75.4 & 69.0 & 76.8 & 68.7 & 76.5 & 67.7 & 76.0 & 67.1 & 74.8 & 62.4 & 67.6 & 66.9 & 73.2 \\
AIDE~\cite{Yan_2025_AIDE} & \textbf{89.6} & \textbf{95.3} & 86.6 & \underline{94.3} & 87.2 & 94.9 & \underline{88.2} & \underline{95.3} & \textbf{88.2} & \underline{95.2} & 85.5 & 93.6 & 85.5 & 93.6 & 85.7 & 93.5 & 87.1 & 94.5 \\
SAFE~\cite{Li_2025_SAFE} & 63.1 & 56.6 & 56.8 & 58.9 & 57.4 & 59.1 & 58.0 & 60.1 & 57.2 & 59.5 & 55.3 & 56.7 & 54.2 & 55.2 & 60.0 & 61.7 & 57.7 & 58.5 \\
Effort~\cite{Yan_2025_Effort} & 66.2 & 65.2 & 69.5 & 73.5 & 69.1 & 74.7 & 71.7 & 75.1 & 72.8 & 73.5 & 70.3 & 73.6 & 69.5 & 71.4 & 66.9 & 72.6 & 69.5 & 72.5 \\
BSF~\cite{Zhou2026BSF} & 53.8 & 55.1 & 57.2 & 62.5 & 57.4 & 62.3 & 57.7 & 62.2 & 57.4 & 63.4 & 56.8 & 61.8 & 56.2 & 61.4 & 54.8 & 61.4 & 56.4 & 61.3 \\
DFFreq~\cite{Yan_2026_DFFreq} & 68.0 & 65.4 & 65.7 & 71.4 & 65.0 & 71.1 & 65.8 & 71.7 & 65.9 & 71.6 & 65.0 & 70.9 & 64.6 & 70.3 & 67.5 & 73.5 & 65.9 & 70.7 \\
LTD~\cite{Yang_2026_LTD} & 89.1 & 93.9 & \underline{88.1} & 93.9 & \underline{87.9} & \underline{95.4} & 86.8 & 94.7 & 86.4 & 94.6 & \underline{87.6} & \textbf{96.7} & \textbf{88.9} & \underline{94.6} & \underline{86.8} & \underline{95.2} & \underline{87.7} & \underline{94.9} \\
IAPL~\cite{Li_2026_IAPL} & 87.6 & 90.1 & 83.5 & 89.6 & 83.5 & 90.2 & 83.8 & 89.9 & 82.8 & 89.4 & 82.1 & 88.4 & 83.7 & 91.3 & 77.3 & 82.8 & 83.0 & 89.0 \\
PPM-CLIP~\cite{Wang_2026_PPMCLIP} & 68.9 & 84.0 & 68.4 & 74.8 & 67.9 & 73.8 & 70.3 & 77.6 & 70.5 & 77.8 & 67.6 & 73.9 & 66.4 & 73.2 & 72.7 & 81.9 & 69.1 & 77.1 \\
\midrule
\rowcolor{suppours}
\textbf{MAFDNet (Ours)} & \underline{89.2} & \underline{94.8} & \textbf{88.5} & \textbf{95.9} & \textbf{88.2} & \textbf{95.7} & \textbf{88.4} & \textbf{95.8} & \underline{87.8} & \textbf{95.6} & \textbf{88.1} & \underline{95.7} & \underline{88.0} & \textbf{95.5} & \textbf{87.8} & \textbf{95.5} & \textbf{88.3} & \textbf{95.6} \\
\bottomrule
\end{tabularx}
\end{table*}

\begin{figure*}[!t]
\centering
\includegraphics[width=\textwidth]{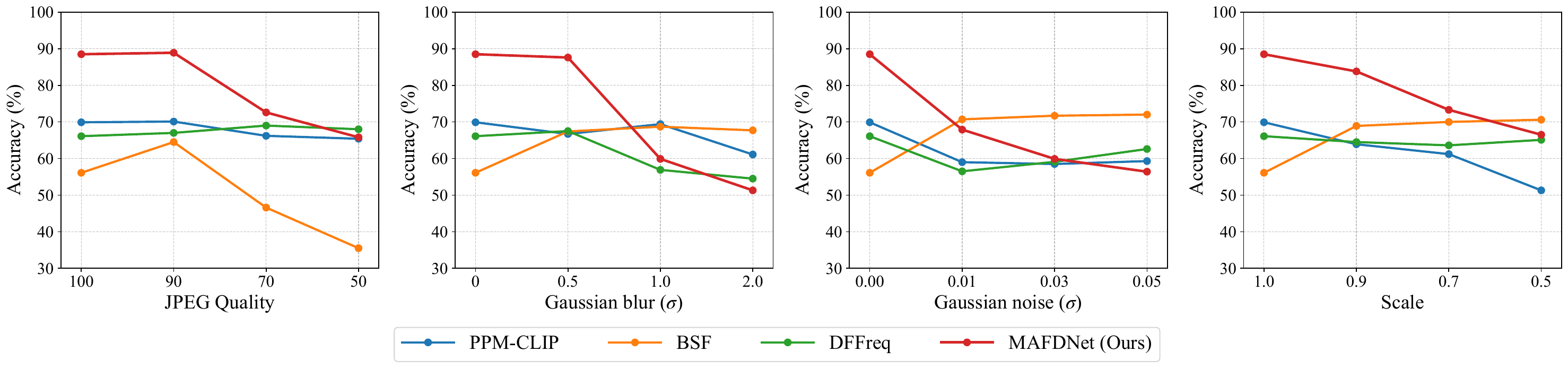}
\caption{Detection accuracy under JPEG compression, Gaussian blur, Gaussian noise, and image downscaling. The horizontal axes indicate JPEG quality, Gaussian standard deviation $\sigma$, and scale factor. MAFDNet retains an advantage under several mild-to-moderate degradations, while stronger blur and noise substantially reduce its accuracy.}
\label{fig:supp_robustness}
\end{figure*}

\begin{figure*}[!t]
\centering
\includegraphics[width=\textwidth]{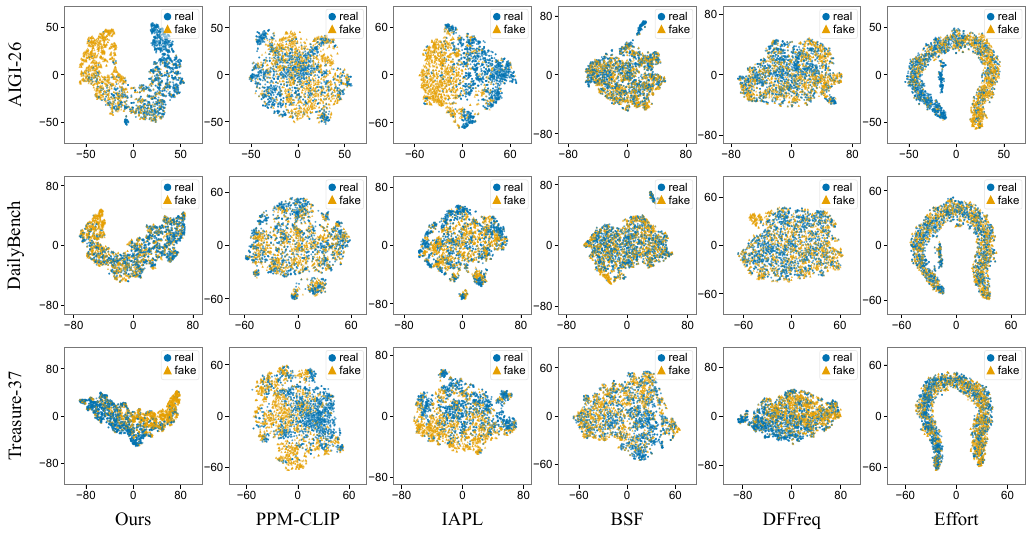}
\caption{t-SNE visualizations of six detectors on AIGI-26, DailyBench, and Treasure-37 (rows).}
\label{fig:supp_tsne}
\end{figure*}

\section{Extended Results and Analysis}
\label{sec:supp_results}

\subsection{Detection across Image Sizes}
\label{sec:supp_scales}
Table~\ref{tab:supp_scale_results} reports Acc and AP for the 16 comparison methods and MAFDNet using their AIGI-26-trained weights. We group the AIGI-26 test images by the eight sizes listed in Table~\ref{tab:supp_resolutions} and evaluate each group separately. These sizes describe the dataset images before detector-specific cropping, rather than different network input sizes. MAFDNet ranks among the top two in both metrics in every group, achieving the highest Acc in five groups and the highest AP in six, including one tie. Its Acc ranges from 87.8\% to 89.2\% and AP from 94.8\% to 95.9\%, showing consistent detection across square, portrait, and landscape formats. The 1.4-point Acc spread is smaller than those of AIDE and LTD, suggesting less variation across the evaluated size groups. MAFDNet also remains competitive on the widest format, where several alternatives lose accuracy. This consistency indicates that its advantage is not confined to square images, supporting the applicability of the combined spatial and spectral evidence across the image sizes represented in AIGI-26.

\subsection{Generator-Wise Detection on AIGI-26}
Table~\ref{tab:supp_generator_results} reports the complete generator-wise results for BSF, DFFreq, LTD, IAPL, PPM-CLIP, and MAFDNet using AIGI-26-trained models. MAFDNet attains the highest Acc on 18 of the 26 generators, including one tie, and ranks among the top two distinct scores on 21 generators for Acc and 25 for AP. Its AP ranges from 83.5\% to 99.9\% and leads on 18 generators, indicating consistently strong ranking performance across the collection. This suggests sensitivity to diverse artifacts. LTD remains competitive in AP and leads on several generators, including SANA-1.5 and OmniGen2. Together, these results support the effectiveness of combining spatial evidence with spectral statistics across diverse generation sources, while showing that the relative strengths of detectors depend on the generator.

\subsection{Generator-Wise Cross-Dataset Evaluation}
Tables~\ref{tab:supp_daily_results} and~\ref{tab:supp_treasure_results} extend the cross-dataset comparison to all seven evaluated DailyBench generators and all 37 Treasure-37 generators. We report Acc and AP for the same six detectors using their AIGI-26-trained weights. MAFDNet achieves the highest Acc on all seven DailyBench generators and the highest AP on five, indicating effective transfer to several independent generation sources. This supports a transferable authenticity decision boundary. The contrast between leading Acc and lower AP on DailyBench's Nano Banana 2 subset also shows that thresholded recognition and score ranking need not agree. Considering both metrics therefore gives a more complete account of transfer than either alone. On Treasure-37, the relative strengths vary across generators. MAFDNet performs strongly on sources such as Sora-image and Wan2.5-T2I-Preview, while Midjourney v6.1 and CogView4 remain challenging. These results support transfer beyond AIGI-26 while identifying generator-specific variation that aggregate scores alone do not reveal.

\subsection{Robustness to Image Degradations}
\label{sec:supp_robustness}
\noindent\textbf{Image processing.}
Each degradation is applied independently to the decoded RGB image. JPEG tests re-encode images at qualities 90, 70, and 50 with fixed 4:4:4 chroma sampling. Gaussian blur uses standard deviations $\sigma\in\{0.5,1.0,2.0\}$ pixels with reflection padding. Gaussian noise adds independent zero-mean perturbations to RGB values normalized to $[0,1]$, with $\sigma\in\{0.01,0.03,0.05\}$, followed by clipping and conversion back to 8-bit values. Reproducible random noise is shared across severity levels and scaled by $\sigma$. Downscaling uses bicubic interpolation with factors 0.9, 0.7, and 0.5, without restoring the original dimensions.

\noindent\textbf{Results and analysis.}
Figure~\ref{fig:supp_robustness} shows that JPEG compression, Gaussian noise, blur, and downscaling affect nearly all evaluated detectors. MAFDNet maintains the highest Acc under mild-to-moderate JPEG compression and downscaling, as well as mild blur, despite performance degradation as perturbations intensify. These results indicate that its complementary forensic cues remain effective under several common forms of moderate image-quality loss.

Gaussian blur particularly affects MAFDNet, reducing Acc from approximately 88\% without blur to 60\% at $\sigma=1.0$ and near chance at $\sigma=2.0$. This is consistent with smoothing suppressing fine textures and local dependencies emphasized by fixed forensic operators, while attenuating high-frequency spectral evidence. Both branches thus remain vulnerable to severe detail loss.

\subsection{Feature Visualization}
\label{sec:supp_tsne}
\noindent\textbf{Visualization procedure.}
For each dataset, we use a shared subset of 2,600 test images, comprising 1,300 real and 1,300 fake images, for all six detectors. Features are extracted in evaluation mode with each detector's preprocessing. We use the inputs to the classification layers for MAFDNet, IAPL, BSF, DFFreq, and Effort, and the normalized image embeddings used for image--text scoring in PPM-CLIP. 

PCA reduces each feature set to 50 dimensions before two-dimensional t-SNE~\cite{Maaten_2008_TSNE}, with PCA initialization and reproducible randomness. Each method--dataset pair is fitted separately, without class labels, which are used only to mark the plotted points.

\noindent\textbf{Results and analysis.}
Figure~\ref{fig:supp_tsne} shows clearer class-associated regions for MAFDNet and IAPL on AIGI-26, whereas PPM-CLIP, BSF, and DFFreq exhibit greater mixing. Effort shows a curved distribution with substantial class overlap. On DailyBench, all methods overlap more, although MAFDNet retains a locally concentrated fake region. On Treasure-37, MAFDNet preserves a fake-dominated region at one end of its distribution, while PPM-CLIP and IAPL also retain class-associated regions. These patterns suggest partially transferable evidence alongside increased ambiguity under distribution shifts. The projections provide qualitative context, rather than a direct measure of classification performance or comparable distances between panels.

\clearpage
\begin{table*}[!t]
\begingroup
\renewenvironment{table*}[1][]{}{}
\begin{table*}[!t]
\centering
\caption{Generator-wise Acc and AP (\%) on the AIGI-26 test set. Results for six detectors are arranged in three panels covering all 26 generators. \textbf{Bold} and \underline{underlined} values denote the best and second-best distinct scores within each generator--metric column, including ties. Generator names follow Table~\ref{tab:supp_generators}.}
\label{tab:supp_generator_results}
\fontsize{8.5}{10.5}\selectfont
\setlength{\tabcolsep}{1.5pt}
\renewcommand{\arraystretch}{1.18}
\begin{tabularx}{\textwidth}{@{}p{72pt}*{18}{>{\centering\arraybackslash}X}@{}}
\toprule
\rowcolor{suppheader}
\textbf{Method} & \multicolumn{2}{c}{\shortstack{Boogu\\0.1 Turbo}} & \multicolumn{2}{c}{\shortstack{ERNIE\\Image}} & \multicolumn{2}{c}{\shortstack{FLUX.2\\klein}} & \multicolumn{2}{c}{\shortstack{Gemini\\2.0 Flash}} & \multicolumn{2}{c}{\shortstack{GLM\\Image}} & \multicolumn{2}{c}{\shortstack{GPT-Image\\1.5}} & \multicolumn{2}{c}{\shortstack{GPT-Image\\2}} & \multicolumn{2}{c}{\shortstack{HiDream\\O1}} & \multicolumn{2}{c}{\shortstack{Hunyuan\\Image 3.0}} \\
\cmidrule(lr){2-3}\cmidrule(lr){4-5}\cmidrule(lr){6-7}\cmidrule(lr){8-9}\cmidrule(lr){10-11}\cmidrule(lr){12-13}\cmidrule(lr){14-15}\cmidrule(lr){16-17}\cmidrule(lr){18-19}
 & Acc & AP & Acc & AP & Acc & AP & Acc & AP & Acc & AP & Acc & AP & Acc & AP & Acc & AP & Acc & AP \\
\midrule
BSF~\cite{Zhou2026BSF} & 64.0 & 76.7 & 50.1 & 49.4 & 48.9 & 50.1 & 55.5 & 68.1 & 63.5 & 79.8 & 62.8 & 66.9 & 62.8 & 69.0 & 47.3 & 44.3 & 65.1 & 65.0 \\
DFFreq~\cite{Yan_2026_DFFreq} & 66.2 & 62.6 & 64.4 & 61.6 & 59.7 & 56.5 & 72.3 & 82.9 & 68.9 & 69.4 & 74.3 & 81.2 & 71.9 & 78.9 & 67.6 & 72.1 & 63.4 & 64.8 \\
LTD~\cite{Yang_2026_LTD} & \underline{89.4} & \underline{94.7} & \textbf{80.8} & \textbf{92.1} & \underline{78.8} & \textbf{87.4} & \underline{92.3} & \underline{95.4} & \underline{92.1} & \underline{95.8} & \underline{91.6} & 96.2 & \underline{91.6} & 95.2 & 78.3 & \underline{92.9} & \underline{92.6} & 96.0 \\
IAPL~\cite{Li_2026_IAPL} & 76.6 & 82.6 & \underline{76.7} & 80.6 & 74.0 & 73.3 & 84.8 & 92.8 & 88.0 & 92.3 & \underline{91.6} & \underline{96.4} & 89.5 & \underline{95.6} & \underline{82.2} & 89.4 & \textbf{94.5} & \underline{98.6} \\
PPM-CLIP~\cite{Wang_2026_PPMCLIP} & 77.7 & 90.7 & 63.8 & 70.6 & 58.1 & 62.5 & 73.3 & 82.3 & 85.1 & 91.3 & 81.1 & 91.0 & 82.1 & 93.5 & 53.1 & 54.0 & 85.0 & 93.9 \\
\rowcolor{suppours}
\textbf{MAFDNet (Ours)} & \textbf{92.3} & \textbf{97.1} & 76.0 & \underline{83.5} & \textbf{79.6} & \underline{87.0} & \textbf{95.5} & \textbf{99.1} & \textbf{97.5} & \textbf{99.8} & \textbf{98.0} & \textbf{99.9} & \textbf{96.8} & \textbf{99.6} & \textbf{84.2} & \textbf{93.1} & \textbf{94.5} & \textbf{98.9} \\
\bottomrule
\end{tabularx}
\par\vspace{7pt}
\begin{tabularx}{\textwidth}{@{}p{72pt}*{18}{>{\centering\arraybackslash}X}@{}}
\toprule
\rowcolor{suppheader}
\textbf{Method} & \multicolumn{2}{c}{\shortstack{Imagen 4\\Ultra}} & \multicolumn{2}{c}{\shortstack{Janus\\Pro}} & \multicolumn{2}{c}{\shortstack{Krea 2\\Turbo}} & \multicolumn{2}{c}{\shortstack{LongCat\\Image}} & \multicolumn{2}{c}{\shortstack{Lumina\\Image 2.0}} & \multicolumn{2}{c}{\shortstack{Mage-Flow\\Turbo}} & \multicolumn{2}{c}{\shortstack{Midjourney\\V7}} & \multicolumn{2}{c}{\shortstack{Nano\\Banana 2}} & \multicolumn{2}{c}{\shortstack{Nano\\Banana Pro}} \\
\cmidrule(lr){2-3}\cmidrule(lr){4-5}\cmidrule(lr){6-7}\cmidrule(lr){8-9}\cmidrule(lr){10-11}\cmidrule(lr){12-13}\cmidrule(lr){14-15}\cmidrule(lr){16-17}\cmidrule(lr){18-19}
 & Acc & AP & Acc & AP & Acc & AP & Acc & AP & Acc & AP & Acc & AP & Acc & AP & Acc & AP & Acc & AP \\
\midrule
BSF~\cite{Zhou2026BSF} & 61.2 & 72.9 & 50.6 & 47.6 & 52.7 & 56.2 & 54.1 & 55.8 & 47.6 & 51.7 & 73.8 & 85.9 & 58.9 & 57.6 & 59.2 & 64.4 & 53.5 & 55.6 \\
DFFreq~\cite{Yan_2026_DFFreq} & 73.1 & 79.9 & 70.7 & 79.1 & 60.2 & 62.8 & 64.4 & 67.6 & 57.7 & 59.9 & 70.8 & 64.4 & 59.9 & 62.5 & 74.2 & 81.0 & 62.6 & 65.9 \\
LTD~\cite{Yang_2026_LTD} & 92.7 & 96.4 & \underline{80.1} & \underline{93.4} & \underline{80.8} & \underline{84.2} & \textbf{86.1} & \textbf{94.4} & \textbf{84.5} & \textbf{94.1} & \underline{92.4} & 95.4 & \textbf{85.0} & \underline{90.1} & \underline{91.5} & \underline{96.1} & \underline{90.8} & \underline{94.7} \\
IAPL~\cite{Li_2026_IAPL} & \underline{93.0} & \underline{97.8} & \underline{80.1} & 91.2 & 67.1 & 68.0 & \underline{85.5} & 89.4 & \underline{81.3} & 83.4 & 92.0 & \underline{95.5} & 76.8 & 86.4 & 90.3 & 95.8 & 76.9 & 86.5 \\
PPM-CLIP~\cite{Wang_2026_PPMCLIP} & 80.5 & 91.0 & 59.0 & 65.6 & 66.9 & 72.5 & 64.2 & 70.6 & 60.2 & 63.1 & 89.6 & 94.7 & 62.2 & 67.1 & 74.2 & 83.0 & 61.5 & 68.9 \\
\rowcolor{suppours}
\textbf{MAFDNet (Ours)} & \textbf{96.3} & \textbf{99.3} & \textbf{87.8} & \textbf{98.6} & \textbf{88.7} & \textbf{96.1} & 85.1 & \underline{93.4} & 79.9 & \underline{89.6} & \textbf{93.0} & \textbf{97.8} & \underline{84.7} & \textbf{95.2} & \textbf{97.0} & \textbf{99.5} & \textbf{91.4} & \textbf{97.1} \\
\bottomrule
\end{tabularx}
\par\vspace{7pt}
\begin{tabularx}{\textwidth}{@{}p{72pt}*{16}{>{\centering\arraybackslash}X}@{}}
\toprule
\rowcolor{suppheader}
\textbf{Method} & \multicolumn{2}{c}{\shortstack{OmniGen2}} & \multicolumn{2}{c}{\shortstack{Qwen\\Image 3.0}} & \multicolumn{2}{c}{\shortstack{SANA\\1.5}} & \multicolumn{2}{c}{\shortstack{SD\\3.5-L}} & \multicolumn{2}{c}{\shortstack{Seedream\\4.0}} & \multicolumn{2}{c}{\shortstack{Seedream\\4.5}} & \multicolumn{2}{c}{\shortstack{SeFi\\Turbo}} & \multicolumn{2}{c}{\shortstack{Z-Image\\Turbo}} \\
\cmidrule(lr){2-3}\cmidrule(lr){4-5}\cmidrule(lr){6-7}\cmidrule(lr){8-9}\cmidrule(lr){10-11}\cmidrule(lr){12-13}\cmidrule(lr){14-15}\cmidrule(lr){16-17}
 & Acc & AP & Acc & AP & Acc & AP & Acc & AP & Acc & AP & Acc & AP & Acc & AP & Acc & AP \\
\midrule
BSF~\cite{Zhou2026BSF} & 52.3 & 56.0 & 52.1 & 55.1 & 49.4 & 49.7 & 48.2 & 48.5 & 62.1 & 74.4 & 65.4 & 82.1 & 48.9 & 52.5 & 50.0 & 57.5 \\
DFFreq~\cite{Yan_2026_DFFreq} & 58.6 & 60.6 & 69.8 & 73.0 & 68.0 & 73.0 & 55.1 & 55.5 & 69.6 & 75.2 & 70.5 & 76.6 & 62.9 & 65.3 & 61.7 & 62.8 \\
LTD~\cite{Yang_2026_LTD} & \textbf{85.5} & \textbf{93.9} & \underline{80.6} & \underline{93.3} & \underline{82.1} & \textbf{95.3} & \textbf{83.8} & \textbf{91.1} & \underline{92.7} & \underline{94.6} & \underline{92.9} & 96.0 & \textbf{87.2} & \textbf{93.5} & \underline{87.7} & \underline{93.1} \\
IAPL~\cite{Li_2026_IAPL} & 79.2 & 81.0 & 73.0 & 76.2 & \textbf{84.8} & \underline{92.1} & 73.0 & 75.3 & 88.1 & 94.5 & 91.8 & \underline{96.6} & \underline{82.1} & 86.8 & 80.0 & 82.7 \\
PPM-CLIP~\cite{Wang_2026_PPMCLIP} & 64.8 & 70.6 & 62.9 & 72.8 & 70.5 & 76.7 & 62.9 & 66.6 & 82.2 & 90.6 & 79.6 & 92.5 & 57.9 & 59.1 & 59.0 & 60.4 \\
\rowcolor{suppours}
\textbf{MAFDNet (Ours)} & \underline{80.5} & \underline{89.3} & \textbf{87.2} & \textbf{97.0} & 75.1 & 83.9 & \underline{79.5} & \underline{86.9} & \textbf{93.8} & \textbf{98.7} & \textbf{96.4} & \textbf{99.4} & 79.0 & \underline{89.8} & \textbf{91.4} & \textbf{97.3} \\
\bottomrule
\end{tabularx}
\end{table*}

\vspace{10pt}
\begin{table*}[!t]
\centering
\caption{Generator-wise Acc and AP (\%) on DailyBench using AIGI-26-trained models. \textbf{Bold} and \underline{underlined} values indicate the best and second-best distinct scores in each generator--metric column, including ties.}
\label{tab:supp_daily_results}
\fontsize{8.5}{10.5}\selectfont
\setlength{\tabcolsep}{1.5pt}
\renewcommand{\arraystretch}{1.18}
\begin{tabularx}{\textwidth}{@{}p{72pt}*{14}{>{\centering\arraybackslash}X}@{}}
\toprule
\rowcolor{suppheader}
\textbf{Method} & \multicolumn{2}{c}{\shortstack{FLUX.1}} & \multicolumn{2}{c}{\shortstack{FLUX.2}} & \multicolumn{2}{c}{\shortstack{GPT-Image\\2}} & \multicolumn{2}{c}{\shortstack{Nano\\Banana 2}} & \multicolumn{2}{c}{\shortstack{Qwen-Image\\2512}} & \multicolumn{2}{c}{\shortstack{SD\\3.5 Large}} & \multicolumn{2}{c}{\shortstack{Z-Image}} \\
\cmidrule(lr){2-3}\cmidrule(lr){4-5}\cmidrule(lr){6-7}\cmidrule(lr){8-9}\cmidrule(lr){10-11}\cmidrule(lr){12-13}\cmidrule(lr){14-15}
 & Acc & AP & Acc & AP & Acc & AP & Acc & AP & Acc & AP & Acc & AP & Acc & AP \\
\midrule
BSF~\cite{Zhou2026BSF} & 44.7 & 45.1 & 46.1 & 48.9 & 68.8 & \underline{86.4} & 46.2 & 41.1 & 46.1 & 47.1 & 48.7 & 49.8 & 46.7 & 48.5 \\
DFFreq~\cite{Yan_2026_DFFreq} & \underline{59.5} & 61.3 & 57.6 & 60.0 & \underline{69.3} & 72.1 & 60.9 & 57.7 & \underline{60.6} & 63.2 & 56.7 & 59.2 & 57.7 & 62.0 \\
LTD~\cite{Yang_2026_LTD} & 53.7 & \underline{71.4} & 52.5 & 57.7 & 46.7 & 38.1 & 51.6 & \underline{64.3} & 52.7 & 64.5 & 53.7 & 60.9 & 52.6 & 61.7 \\
IAPL~\cite{Li_2026_IAPL} & 57.8 & 68.6 & \underline{59.5} & \underline{61.9} & 59.7 & 62.3 & \underline{61.0} & \textbf{72.1} & 59.4 & \underline{72.4} & \underline{60.3} & \underline{71.0} & \underline{59.3} & \textbf{68.0} \\
PPM-CLIP~\cite{Wang_2026_PPMCLIP} & 53.0 & 56.4 & 51.6 & 51.1 & 54.6 & 61.7 & 53.1 & 52.1 & 54.1 & 59.1 & 54.2 & 56.5 & 51.4 & 51.5 \\
\rowcolor{suppours}
\textbf{MAFDNet (Ours)} & \textbf{74.0} & \textbf{80.7} & \textbf{59.8} & \textbf{63.0} & \textbf{79.9} & \textbf{90.6} & \textbf{62.8} & 49.4 & \textbf{81.6} & \textbf{90.7} & \textbf{68.5} & \textbf{72.7} & \textbf{62.4} & \underline{66.0} \\
\bottomrule
\end{tabularx}
\end{table*}

\endgroup
\end{table*}
\clearpage
\begin{table*}[!t]
\centering
\caption{Generator-wise Acc and AP (\%) on Treasure-37 using AIGI-26-trained models. \textbf{Bold} and \underline{underlined} values indicate the best and second-best distinct scores in each generator--metric column, including ties.}
\label{tab:supp_treasure_results}
\fontsize{8.5}{10.5}\selectfont
\setlength{\tabcolsep}{1.5pt}
\renewcommand{\arraystretch}{1.18}
\begin{tabularx}{\textwidth}{@{}p{72pt}*{18}{>{\centering\arraybackslash}X}@{}}
\toprule
\rowcolor{suppheader}
\textbf{Method} & \multicolumn{2}{c}{\shortstack{BAGEL}} & \multicolumn{2}{c}{\shortstack{BRIA\\3.2}} & \multicolumn{2}{c}{\shortstack{CogView3\\plus}} & \multicolumn{2}{c}{\shortstack{CogView4}} & \multicolumn{2}{c}{\shortstack{Seedream\\3.0}} & \multicolumn{2}{c}{\shortstack{Seedream\\4.0}} & \multicolumn{2}{c}{\shortstack{FLUX.2}} & \multicolumn{2}{c}{\shortstack{GPT-Image\\1.5}} & \multicolumn{2}{c}{\shortstack{HiDream\\I1-Dev}} \\
\cmidrule(lr){2-3}\cmidrule(lr){4-5}\cmidrule(lr){6-7}\cmidrule(lr){8-9}\cmidrule(lr){10-11}\cmidrule(lr){12-13}\cmidrule(lr){14-15}\cmidrule(lr){16-17}\cmidrule(lr){18-19}
 & Acc & AP & Acc & AP & Acc & AP & Acc & AP & Acc & AP & Acc & AP & Acc & AP & Acc & AP & Acc & AP \\
\midrule
BSF~\cite{Zhou2026BSF} & 57.8 & 63.4 & 61.8 & 68.2 & 58.3 & 59.6 & 48.6 & 40.8 & 67.4 & 64.4 & 61.3 & 60.3 & \underline{64.8} & 69.5 & 65.6 & 72.6 & 60.8 & 69.5 \\
DFFreq~\cite{Yan_2026_DFFreq} & 61.8 & 65.1 & 60.6 & 60.5 & \underline{65.5} & 71.0 & 59.7 & 58.4 & 63.0 & 62.9 & \underline{64.9} & 60.9 & 58.0 & 53.0 & \textbf{71.5} & 69.7 & 60.1 & 51.9 \\
LTD~\cite{Yang_2026_LTD} & \underline{66.0} & \underline{79.6} & 71.8 & 87.8 & 65.0 & 80.5 & \textbf{68.8} & \textbf{92.5} & 65.5 & 75.6 & 52.4 & 58.3 & 60.1 & \underline{71.9} & 63.0 & 71.6 & 64.7 & 82.3 \\
IAPL~\cite{Li_2026_IAPL} & 63.2 & 72.8 & 69.3 & \underline{89.5} & \textbf{67.5} & \textbf{86.2} & \underline{67.5} & \underline{80.5} & 62.8 & 70.5 & 64.1 & 71.7 & 63.4 & 64.0 & 67.3 & 67.5 & 62.5 & 66.8 \\
PPM-CLIP~\cite{Wang_2026_PPMCLIP} & 62.7 & \textbf{80.1} & \textbf{73.8} & \textbf{92.2} & 62.8 & \underline{85.8} & 64.0 & 78.1 & \underline{68.2} & \underline{83.3} & 64.8 & \underline{81.5} & \textbf{66.4} & \textbf{83.3} & 70.0 & \underline{87.2} & \underline{71.3} & \underline{84.1} \\
\rowcolor{suppours}
\textbf{MAFDNet (Ours)} & \textbf{68.9} & 76.7 & \underline{71.9} & 82.6 & 64.6 & 76.5 & 55.1 & 43.9 & \textbf{74.4} & \textbf{94.9} & \textbf{73.2} & \textbf{97.7} & 64.3 & 70.9 & \underline{70.5} & \textbf{96.0} & \textbf{72.3} & \textbf{98.8} \\
\bottomrule
\end{tabularx}
\par\vspace{7pt}
\begin{tabularx}{\textwidth}{@{}p{72pt}*{18}{>{\centering\arraybackslash}X}@{}}
\toprule
\rowcolor{suppheader}
\textbf{Method} & \multicolumn{2}{c}{\shortstack{Hunyuan\\DiT}} & \multicolumn{2}{c}{\shortstack{Hunyuan\\Image 3.0}} & \multicolumn{2}{c}{\shortstack{Imagen\\4}} & \multicolumn{2}{c}{\shortstack{Infinity\\2B}} & \multicolumn{2}{c}{\shortstack{Janus-Pro\\7B}} & \multicolumn{2}{c}{\shortstack{Kolors}} & \multicolumn{2}{c}{\shortstack{LlamaGen}} & \multicolumn{2}{c}{\shortstack{LongCat\\Image}} & \multicolumn{2}{c}{\shortstack{Lumina\\Image 2.0}} \\
\cmidrule(lr){2-3}\cmidrule(lr){4-5}\cmidrule(lr){6-7}\cmidrule(lr){8-9}\cmidrule(lr){10-11}\cmidrule(lr){12-13}\cmidrule(lr){14-15}\cmidrule(lr){16-17}\cmidrule(lr){18-19}
 & Acc & AP & Acc & AP & Acc & AP & Acc & AP & Acc & AP & Acc & AP & Acc & AP & Acc & AP & Acc & AP \\
\midrule
BSF~\cite{Zhou2026BSF} & 61.1 & 71.6 & 67.9 & 67.8 & 64.2 & 72.9 & 64.4 & 65.7 & 66.3 & 69.9 & 63.0 & 69.4 & 64.1 & 68.4 & 67.7 & 81.3 & 65.5 & 78.4 \\
DFFreq~\cite{Yan_2026_DFFreq} & 63.2 & 65.6 & 61.1 & 57.5 & 62.2 & 62.2 & \textbf{65.5} & 68.1 & 66.0 & 63.9 & 65.5 & 65.1 & 53.1 & 51.6 & 63.0 & 65.7 & 58.3 & 58.1 \\
LTD~\cite{Yang_2026_LTD} & 68.3 & 87.9 & 67.1 & 79.0 & 66.5 & 78.1 & 65.2 & \underline{82.1} & 64.2 & 76.9 & \textbf{74.2} & 90.9 & 67.3 & 77.0 & 64.4 & 69.8 & \underline{68.9} & 78.9 \\
IAPL~\cite{Li_2026_IAPL} & \textbf{71.3} & \underline{89.0} & \underline{68.0} & 80.2 & \underline{66.7} & 71.6 & \underline{65.3} & \textbf{83.0} & 65.8 & \underline{86.5} & 70.3 & \underline{92.5} & \underline{71.2} & \underline{88.5} & \underline{68.5} & 81.0 & 67.0 & 78.1 \\
PPM-CLIP~\cite{Wang_2026_PPMCLIP} & \underline{70.1} & \textbf{92.3} & 66.6 & \underline{89.6} & 64.0 & \underline{82.5} & 64.8 & 79.2 & \underline{68.5} & \textbf{88.7} & \underline{72.7} & \textbf{92.6} & \textbf{74.3} & \textbf{92.6} & 68.3 & \underline{84.4} & \textbf{71.8} & \textbf{87.1} \\
\rowcolor{suppours}
\textbf{MAFDNet (Ours)} & 65.6 & 74.7 & \textbf{72.7} & \textbf{91.3} & \textbf{68.4} & \textbf{86.6} & 62.7 & 65.7 & \textbf{71.7} & 76.8 & 65.6 & 81.5 & 68.0 & 75.5 & \textbf{74.4} & \textbf{94.0} & \underline{68.9} & \underline{81.1} \\
\bottomrule
\end{tabularx}
\par\vspace{7pt}
\begin{tabularx}{\textwidth}{@{}p{72pt}*{18}{>{\centering\arraybackslash}X}@{}}
\toprule
\rowcolor{suppheader}
\textbf{Method} & \multicolumn{2}{c}{\shortstack{Midjourney\\v6.1}} & \multicolumn{2}{c}{\shortstack{Midjourney\\v7}} & \multicolumn{2}{c}{\shortstack{Nano\\Banana}} & \multicolumn{2}{c}{\shortstack{Nano\\Banana Pro}} & \multicolumn{2}{c}{\shortstack{NextStep\\1}} & \multicolumn{2}{c}{\shortstack{OmniGen}} & \multicolumn{2}{c}{\shortstack{OmniGen2}} & \multicolumn{2}{c}{\shortstack{Ovis-U1}} & \multicolumn{2}{c}{\shortstack{Playground\\V2}} \\
\cmidrule(lr){2-3}\cmidrule(lr){4-5}\cmidrule(lr){6-7}\cmidrule(lr){8-9}\cmidrule(lr){10-11}\cmidrule(lr){12-13}\cmidrule(lr){14-15}\cmidrule(lr){16-17}\cmidrule(lr){18-19}
 & Acc & AP & Acc & AP & Acc & AP & Acc & AP & Acc & AP & Acc & AP & Acc & AP & Acc & AP & Acc & AP \\
\midrule
BSF~\cite{Zhou2026BSF} & 54.5 & 46.2 & 51.4 & 42.7 & \underline{68.1} & 72.5 & 41.0 & 39.1 & 66.1 & 56.2 & 61.3 & 73.1 & 57.3 & 61.6 & 62.0 & 65.1 & 65.6 & 63.1 \\
DFFreq~\cite{Yan_2026_DFFreq} & 53.6 & 52.7 & 45.1 & 42.5 & 64.4 & 63.5 & 55.0 & 57.4 & 70.1 & 65.4 & 57.8 & 58.7 & 63.9 & 67.5 & 62.0 & 64.4 & 59.0 & 53.5 \\
LTD~\cite{Yang_2026_LTD} & 59.8 & \underline{68.5} & 50.8 & 52.8 & 65.9 & 75.6 & \underline{59.6} & \underline{63.1} & \textbf{71.7} & \textbf{94.8} & 67.4 & 82.1 & \underline{66.4} & 84.2 & 66.5 & \textbf{92.8} & \textbf{72.8} & \underline{90.2} \\
IAPL~\cite{Li_2026_IAPL} & \underline{61.4} & 67.3 & \underline{62.3} & \underline{63.4} & 65.5 & 68.3 & 59.0 & 59.3 & 67.3 & 89.1 & \underline{69.3} & \underline{83.4} & \textbf{67.7} & \underline{84.9} & \underline{69.2} & 90.2 & 70.6 & 89.9 \\
PPM-CLIP~\cite{Wang_2026_PPMCLIP} & \textbf{62.6} & \textbf{72.3} & \textbf{65.5} & \textbf{78.2} & 64.5 & \underline{84.5} & 50.2 & 54.0 & \underline{70.8} & 89.9 & \textbf{73.7} & \textbf{90.0} & 63.5 & \textbf{88.7} & 67.5 & \underline{91.5} & 68.4 & \textbf{90.3} \\
\rowcolor{suppours}
\textbf{MAFDNet (Ours)} & 51.2 & 40.9 & 57.6 & 52.7 & \textbf{70.9} & \textbf{85.7} & \textbf{66.4} & \textbf{68.7} & \textbf{71.7} & \underline{94.2} & 68.2 & 76.2 & \underline{66.4} & 76.2 & \textbf{70.3} & 85.2 & \underline{72.5} & 76.8 \\
\bottomrule
\end{tabularx}
\par\vspace{7pt}
\begin{tabularx}{\textwidth}{@{}p{72pt}*{20}{>{\centering\arraybackslash}X}@{}}
\toprule
\rowcolor{suppheader}
\textbf{Method} & \multicolumn{2}{c}{\shortstack{Playground\\V2.5}} & \multicolumn{2}{c}{\shortstack{Qwen\\Image}} & \multicolumn{2}{c}{\shortstack{SANA\\v1.5}} & \multicolumn{2}{c}{\shortstack{SD3\\medium}} & \multicolumn{2}{c}{\shortstack{Show-o}} & \multicolumn{2}{c}{\shortstack{Show-o2}} & \multicolumn{2}{c}{\shortstack{Sora\\image}} & \multicolumn{2}{c}{\shortstack{Wan2.2\\T2I-Flash}} & \multicolumn{2}{c}{\shortstack{Wan2.5\\Preview}} & \multicolumn{2}{c}{\shortstack{Z-Image\\Turbo}} \\
\cmidrule(lr){2-3}\cmidrule(lr){4-5}\cmidrule(lr){6-7}\cmidrule(lr){8-9}\cmidrule(lr){10-11}\cmidrule(lr){12-13}\cmidrule(lr){14-15}\cmidrule(lr){16-17}\cmidrule(lr){18-19}\cmidrule(lr){20-21}
 & Acc & AP & Acc & AP & Acc & AP & Acc & AP & Acc & AP & Acc & AP & Acc & AP & Acc & AP & Acc & AP & Acc & AP \\
\midrule
BSF~\cite{Zhou2026BSF} & 64.4 & 68.1 & 62.2 & 69.3 & 67.7 & 73.9 & 64.2 & 74.6 & 65.8 & 68.4 & 64.1 & 72.2 & 62.5 & 69.6 & 64.9 & 68.5 & 66.7 & 72.2 & 63.7 & 80.4 \\
DFFreq~\cite{Yan_2026_DFFreq} & 62.5 & 56.9 & \textbf{72.6} & 72.6 & 64.2 & 68.8 & 61.8 & 62.6 & 57.6 & 58.6 & 62.3 & 65.5 & \underline{70.3} & 76.7 & 64.2 & 65.3 & 68.2 & 71.7 & 69.3 & 69.4 \\
LTD~\cite{Yang_2026_LTD} & 65.4 & 89.4 & 71.4 & \underline{90.0} & \underline{68.6} & \underline{92.6} & 68.0 & 79.3 & \textbf{71.7} & \underline{85.1} & 64.8 & 87.4 & \textbf{71.5} & 77.5 & \textbf{71.3} & \underline{87.5} & 65.0 & 69.3 & \underline{71.1} & \underline{80.8} \\
IAPL~\cite{Li_2026_IAPL} & \underline{70.5} & \underline{90.6} & 69.5 & 82.8 & 68.3 & \textbf{92.8} & 68.7 & \underline{81.5} & 67.0 & 83.4 & \underline{69.3} & \textbf{90.0} & 65.2 & 68.9 & 67.7 & 72.4 & \underline{69.8} & 75.2 & 63.2 & 70.3 \\
PPM-CLIP~\cite{Wang_2026_PPMCLIP} & \textbf{71.0} & \textbf{92.5} & 70.5 & 89.6 & \textbf{69.0} & 92.3 & \textbf{69.7} & \textbf{85.5} & \underline{69.7} & \textbf{91.7} & 66.3 & 87.7 & 65.2 & \underline{87.1} & \underline{68.9} & 86.4 & 68.1 & \underline{86.3} & 64.0 & 76.6 \\
\rowcolor{suppours}
\textbf{MAFDNet (Ours)} & 67.8 & 75.5 & \underline{71.7} & \textbf{91.4} & 67.2 & 76.2 & \underline{68.9} & 78.1 & 66.8 & 73.0 & \textbf{72.1} & \underline{89.4} & 70.1 & \textbf{92.6} & 68.6 & \textbf{88.2} & \textbf{70.1} & \textbf{98.6} & \textbf{76.2} & \textbf{88.6} \\
\bottomrule
\end{tabularx}
\end{table*}

\clearpage


\end{document}